\documentclass[10pt,twocolumn,letterpaper]{article}

\usepackage[pagenumbers]{cvpr} %

\usepackage{graphicx}
\usepackage{amsmath}
\usepackage{amssymb}
\usepackage{booktabs}

\usepackage{makecell}
\usepackage{multirow}

\usepackage{mathtools}

\usepackage{algorithm}
\usepackage{listings}
\usepackage[table]{xcolor}
\usepackage{booktabs} %
\usepackage{colortbl}
\usepackage{glossaries}
\usepackage{multirow}
\usepackage[font=small]{caption}

\usepackage{pifont}%
\usepackage{xspace}
\usepackage{placeins}
\usepackage{tikz}
\usepackage{comment}
\usepackage{xspace}
\usepackage{color}
\usepackage{subcaption}

\usepackage[pagebackref,breaklinks,colorlinks]{hyperref}

\usepackage[capitalize]{cleveref}
\crefname{section}{Sec.}{Secs.}
\Crefname{section}{Section}{Sections}
\Crefname{table}{Table}{Tables}
\crefname{table}{Tab.}{Tabs.}

\def\cvprPaperID{***} %
\def\confName{CVPR}
\def\confYear{2023}

\begin{document}

\title{Introspective Deep Metric Learning for Image Retrieval}

\newcommand*\samethanks[1][\value{footnote}]{\footnotemark[#1]}
\author{Wenzhao Zheng\footnotetext{sadfsadfa}\thanks{Equal contribution.}\quad Chengkun Wang\samethanks\quad  Jie Zhou\quad Jiwen Lu\thanks{Corresponding author.} \\
Beijing National Research Center for Information Science and Technology, China \\
Department of Automation, Tsinghua University, China \\
\texttt{\{zhengwz18,wck20\}@mails.tsinghua.edu.cn;} 
\texttt{\{jzhou,lujiwen\}@tsinghua.edu.cn}
}

\maketitle

\begin{abstract}
This paper proposes an introspective deep metric learning (IDML) framework for uncertainty-aware comparisons of images. 
Conventional deep metric learning methods produce confident semantic distances between images regardless of the uncertainty level. 
However, we argue that a good similarity model should consider the semantic discrepancies with caution to better deal with ambiguous images for more robust training.
To achieve this, we propose to represent an image using not only a semantic embedding but also an accompanying uncertainty embedding, which describes the semantic characteristics and ambiguity of an image, respectively.
We further propose an introspective similarity metric to make similarity judgments between images considering both their semantic differences and ambiguities. 
The proposed IDML framework improves the performance of deep metric learning through uncertainty modeling and attains state-of-the-art results on the widely used CUB-200-2011, Cars196, and Stanford Online Products datasets for image retrieval and clustering.
We further provide an in-depth analysis of our framework to demonstrate the effectiveness and reliability of IDML.
Code is available at: \url{https://github.com/wzzheng/IDML}.

\end{abstract} 
\newcommand{\tablevspace}{\vspace{-6mm}} %
\newcommand{\figvspace}{\vspace{-5mm}} %

\section{Introduction}
Learning an effective metric to measure the similarity between data is a long-standing problem in computer vision, which serves as a fundamental step in various downstream tasks, such as face recognition~\cite{schroff2015facenet,guo2020density,yang2019learning}, image retrieval~\cite{babenko2014neural,song2016deep,zheng2021deep} and image classification~\cite{deng2019arcface,shi2019probabilistic}. 
The objective of metric learning is to reduce the distances between positive pairs and enlarge the distances between negative pairs, which has recently powered the rapid developments for both supervised learning~\cite{wang2020cross,zhu2020imbalance} and unsupervised learning~\cite{he2020momentum,chen2021exploring,grill2020bootstrap,chen2020simple}. 

\begin{figure}[!t]
\centering
\includegraphics[width=0.495\textwidth]{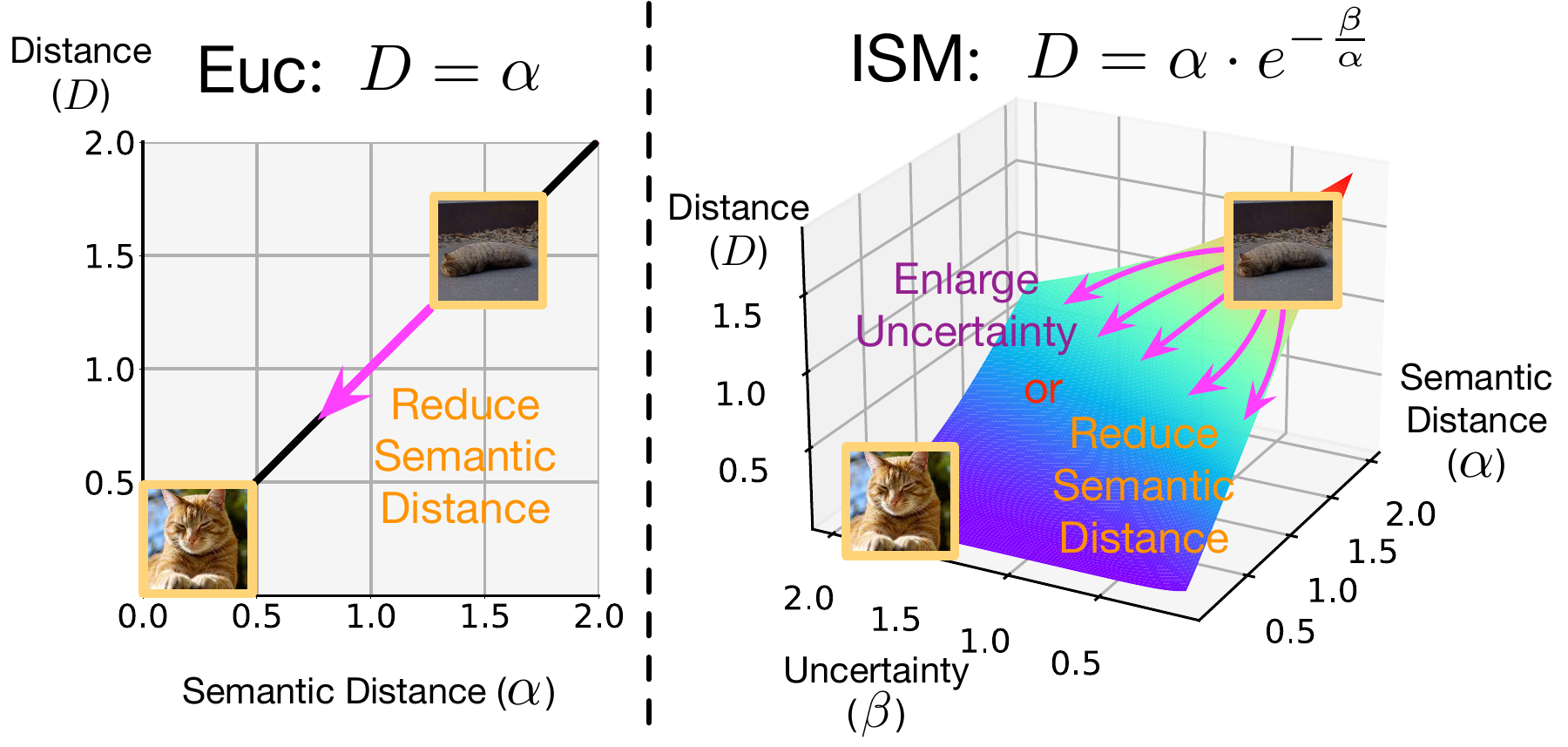}
\vspace{-7mm}
\caption{
For a semantically ambiguous image, conventional DML explicitly reduces its distance with other intraclass images unaware of the uncertainty.
Differently, the proposed introspective similarity metric provides an alternative way to enlarge the uncertainty level to allow confusion in the network.  
} 
\label{fig:motivation}
\figvspace
\end{figure}

Generally, deep metric learning (DML) employs deep neural networks~\cite{simonyan2014very,szegedy2015going,he2016deep} to map an image to a discriminative embedding space~\cite{zhang2017learning}.
Most methods represent images using a deterministic embedding which only describes the characteristic features~\cite{wu2017sampling,schroff2015facenet}.
Nevertheless, when asked to classify a certain image, humans are able to additionally provide the corresponding confidence as well as the semantic features of the image since an image might be ambiguous.
Motivated by this, researchers have proposed a variety of probabilistic embedding methods using distributions to model images in the embedding space~\cite{oh2018modeling,zhang2021point,sun2020view,chun2021probabilistic}. 
They typically use KL-divergence~\cite{hershey2007approximating} or Monte-Carlo-sampling-based~\cite{oh2018modeling} distances to measure the similarity between images.
They regard the variance of the distributions as the uncertainty measure of images, yet they still provide a confident judgment of similarity regardless of the uncertainty.
Specifically, though the variance affects the distribution discrepancy, a larger variance of an image does not necessarily blur its differences from other images.
Given a highly ambiguous image (\emph{e.g.}, an extremely blurred image), we think it is more reasonable to weaken the semantic discrepancies and consider it similar to other images since it literally could be anything.

In this paper, we propose an introspective similarity metric to achieve this and further present an introspective deep metric learning (IDML) framework for image retrieval.
Different from existing methods, we represent an image using a semantic embedding to capture the semantic characteristics and further accompany it with an uncertainty embedding to model the uncertainty. 
An introspective similarity metric then takes as input both embeddings and outputs an uncertainty-aware similarity score, which softens the semantic discrepancies by the degree of uncertainty.
Different from the conventional metric, the proposed introspective metric deems a pair of images similar if they are either semantically similar or ambiguous to judge, as illustrated in Figure~\ref{fig:motivation}.
It provides more flexibility to the training process to avoid the harmful influence of inaccurately labeled data.
To further demonstrate the advantage of the proposed metric for learning with uncertainties, we employ the widely used Mixup~\cite{zhang2018mixup,chou2020remix,verma2019manifold} technique to generate images with large uncertainties and further employ the IDML framework for learning. 
The overall framework of the proposed IDML can be trained efficiently in an end-to-end manner and generally applied to existing deep metric learning methods.
We perform extensive experiments on the CUB-200-2011, Cars196, and Stanford Online Products datasets for image retrieval, which shows that our framework generally improves the performance of existing deep metric learning methods by a large margin and attains state-of-the-art results.
We additionally provide an in-depth analysis of our framework including an ablation study of different components, effects of different augmentations, and qualitative analysis of the learned uncertainty.

\section{Related Work}
\textbf{Deep Metric Learning:}
Deep metric learning aims to construct an effective embedding space to measure the similarity between images
The objective is to decrease intraclass distances and increase interclass distances. 
Most methods~\cite{cakir2019deep,do2019theoretically,wang2014learning,song2016deep,wang2017deep,wang2019multi,yu2019deep,hu2014discriminative,deng2019arcface,kim2020proxy,wang2018cosface,movshovitz2017no,zheng2020structural,zhang2022attributable} employ a discriminative loss to learn the image embeddings.
For example, the commonly used contrastive loss~\cite{hu2014discriminative} pulls embeddings from the same class as close as possible while maintaining a fixed margin between embeddings from different classes. 
Wang~\emph{et al.}~\cite{wang2019multi} further formulated three types of similarities between embeddings and proposed a multi-similarity loss to restrict them. 
The ProxyNCA loss~\cite{movshovitz2017no} generates a proxy for each class and instead constrains the distances between embeddings and proxies. 

In addition to the design of loss functions, various methods have explored effective sampling strategies for better training~\cite{zhang2018mixup,yun2019cutmix,wu2017sampling,harwood2017smart,yuan2017hard,schroff2015facenet,duan2018deep,zheng2020deep}. 
For example, hard negative mining~\cite{harwood2017smart,yuan2017hard,schroff2015facenet} selects challenging negative samples for more efficient learning of the metric. 
To further alleviate the lack of informative training samples, recent works~\cite{duan2018deep,zheng2021hardness} proposed to generate synthetic samples for training.
Also, a variety of data augmentation methods improve the performance by mixing original images for better generalization~\cite{zhang2018mixup,yun2019cutmix}. 
The aforementioned methods employ synthetic images for training, which are can be semantically ambiguous. 
We design an introspective similarity metric to consider the uncertainty and further incorporate them to make similarity judgments.

\textbf{Uncertainty Modeling:}
Uncertainty modeling is widely adopted in natural language processing to model the inherent hierarchies of words~\cite{vilnis2015word,nguyen2017mixture,neelakantan2014efficient}. 
Vilnis~\emph{et al.}~\cite{vilnis2015word} proposed the Gaussian formation in word embedding and Nguyen~\emph{et al.}~\cite{nguyen2017mixture} presented a mixture model to learn multi-sense word embeddings. 
The computer vision field has also benefited from uncertainty modeling due to the natural uncertainty in images caused by various factors such as occlusion and blur~\cite{kendall2017uncertainties,shaw2002signal}.
 Various methods have attempted to model the uncertainty for better robustness and generalization in face recognition~\cite{shi2019probabilistic,chang2020data}, point cloud segmentation~\cite{zhang2021point}, and age estimation~\cite{li2021learning}.

A prevailing method is to model each image as a statistical distribution and regard the variance as the uncertainty measure. 
For example, Oh~\emph{et al.}~\cite{oh2018modeling} employed Gaussian distributions to represent images and used Monte-Carlo sampling to sample several point embeddings from the distributions.
They then imposed a soft contrastive loss on the sampled embeddings to optimize the metric. 
Similar strategy has also been used in pose estimation~\cite{sun2020view}, cross-model retrieval~\cite{chun2021probabilistic} and unsupervised embedding learning~\cite{ye2020probabilistic}. 
However, they still make confident similarity judgments regardless of the uncertainty and a larger variance would not necessarily weaken the semantic discrepancies between samples.
Differently, we propose an introspective metric to measure the similarity between images, which tends to omit the semantic differences between two images given a large uncertainty level.
Also, our method bypasses the optimization of distributions to further increase the robustness.

\section{Proposed Approach}

\subsection{Motivation of an Uncertainty-Aware Metric}
Let $\mathbf{X}$ be an image set with $N$ training samples $\{\mathbf{x}_1, ..., \mathbf{x}_N\}$ and $\mathbf{L}$ be the ground truth label set $\{l_1, ..., l_N\}$. 
Deep metric learning methods aim at learning a mapping to transform each image $\mathbf{x}_i$ to an embedding space, where conventional methods~\cite{hu2014discriminative,movshovitz2017no} use a deterministic vector embedding $\mathbf{y}_i$ to represent an image.
They usually adopt the Euclidean distance as the distance measure:
\begin{eqnarray}\label{equ:euclidean}
D(\mathbf{x}_1, \mathbf{x}_2) = D_{E} (\mathbf{x}_1, \mathbf{x}_2) = ||\mathbf{y}_1 - \mathbf{y}_2||_2,
\end{eqnarray}
where $||\cdot||_2$ denotes the L2-norm.
They then impose various constraints on the pairwise distances to enlarge interclass distances and reduce intraclass distances.

\begin{figure}[t]
\centering
\includegraphics[width=0.495\textwidth]{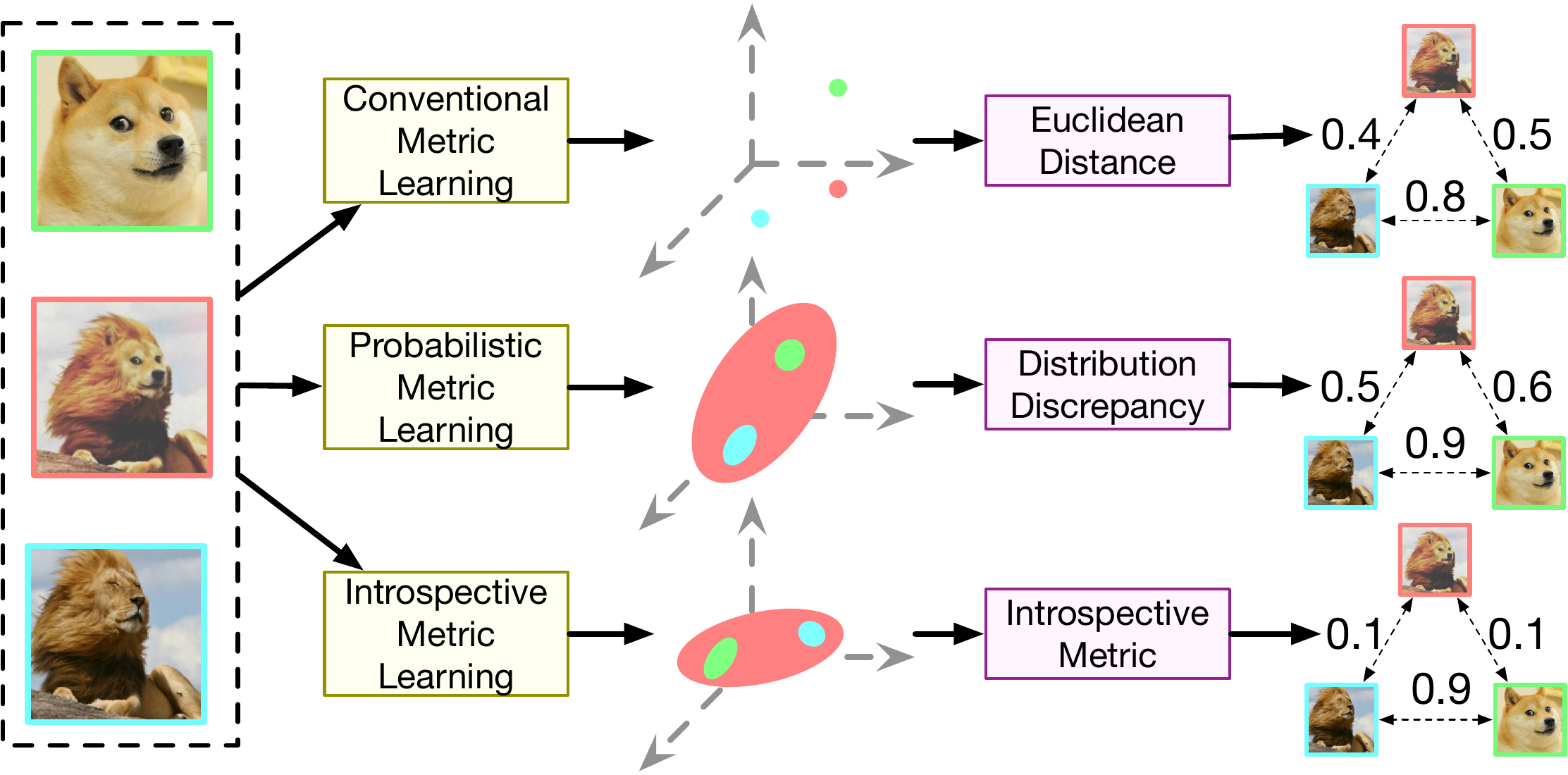}
\vspace{-7mm}
\caption{
Comparisons between different similarity metrics. 
Conventional metric learning and probabilistic metric learning both produce a discriminative distance for a pair of images regardless of the uncertainty level.
Our introspective similarity metric weakens the semantic discrepancies for uncertain pairs.
} 
\label{fig:comparison}
\figvspace
\end{figure}

Conventional DML methods only represent the semantic information in the embedding space, ignoring the possible uncertainty in images.
However, the semantic uncertainty widely exists resulting from low resolution, blur, occlusion, or semantic ambiguity of images, which motivates probabilistic embedding learning approaches~\cite{oh2018modeling,chun2021probabilistic} to model images as statistical distributions $\mathbf{Y}$ in the embedding space. They further use the distribution variance $\mathbf{\sigma}$ to measure the uncertainty of the image in the embedding space.
A widely used method is to employ a Gaussian distribution to describe an image~\cite{oh2018modeling}, \emph{i.e.}, $\mathbf{Y} \sim N(\mathbf{\mu}, \mathbf{\sigma})$, where they use a deep network to learn two vectors $\mathbf{\mu}$ and $\mathbf{\sigma}$ as the mean and variance of the distribution, respectively, assuming each dimension is independent.
They employ distribution divergences~\cite{hershey2007approximating,yu2013kl} or  Monte-Carlo-sampling-based~\cite{oh2018modeling} distances as the similarity metric.
For instance, Hershey~\emph{et al.}~\cite{hershey2007approximating} adopted the KL-divergence to measure the discrepancy of two Gaussian distributions $\mathbf{Y}_1$ and $ \mathbf{Y}_2$:
\begin{eqnarray}
 D_{KL} = 
 -\frac{1}{2} \sum_{k=1}^{d}[log\frac{\sigma_{1,k}^2}{\sigma_{2,k}^2} -  \frac{\sigma_{1,k}^2}{\sigma_{2,k}^2}  -  \frac{(\mu_{1,k}  -   \mu_{2,k})^2}{\sigma_{2,k}^2}   +   1],
\end{eqnarray}
where $d$ denotes the dimension of the Gaussian distributions, $\mathbf{Y}_1 \sim N(\mathbf{\mu}_1, \mathbf{\sigma}_1)$, and $\mathbf{Y}_2 \sim N(\mathbf{\mu}_2, \mathbf{\sigma}_2)$, and $\sigma_{1,k}$ denotes the $k$th component of $\mathbf{\sigma}_1$.

One can find that for two distributions with the same mean, their discrepancy solely depends on the ratio of the variance.
The discrepancy still varies greatly when the variance of one image is large, \emph{i.e.}, of large uncertainty. 
In other words, it still provides confident judgments about the similarity even when uncertain.
However, we argue that a good similarity metric should weaken the semantic discrepancies for uncertain image pairs to allow confusion during training, which has been proven to be useful in knowledge distillation~\cite{hinton2015distilling}. 
This avoids the false pulling of ambiguous pairs to improve the generalization of the learned model.
Figure~\ref{fig:comparison} presents the comparisons between different metrics.

\begin{figure}[t]
\centering
\includegraphics[width=0.495\textwidth]{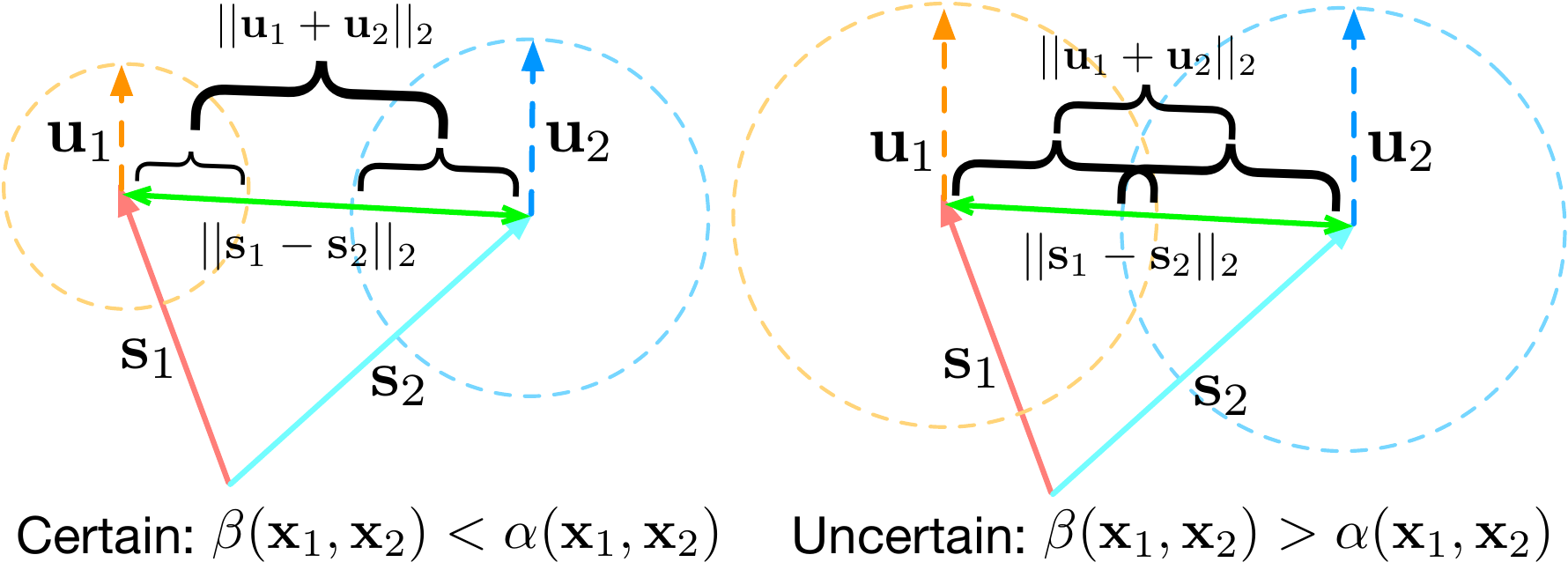}
\vspace{-7mm}
\caption{
Illustration of the proposed uncertain-aware comparison of images.
We consider both the semantic discrepancy $\alpha(\mathbf{x}_1, \mathbf{x}_2) = ||\mathbf{s}_1 - \mathbf{s}_2 ||_2$ and the uncertainty level $\beta(\mathbf{x}_1, \mathbf{x}_2) = ||\mathbf{u}_1 + \mathbf{u}_2 ||_2$ to compute the similarity. 
We deem it uncertain to distinguish a pair when $\beta(\mathbf{x}_1, \mathbf{x}_2) > \alpha(\mathbf{x}_1, \mathbf{x}_2)$.
We only demonstrate the case when $\mathbf{u}_1$ and $\mathbf{u}_2$ align with each other for simplicity.
Practically, we use $||\mathbf{u}_1 + \mathbf{u}_2 ||_2$ instead of $||\mathbf{u}_1||_2 + ||\mathbf{u}_2 ||_2$ to facilitate more capacity.
} 
\label{fig:uncertain}
\figvspace
\end{figure}

\begin{figure*}[t]
\centering
\includegraphics[width=0.98\textwidth]{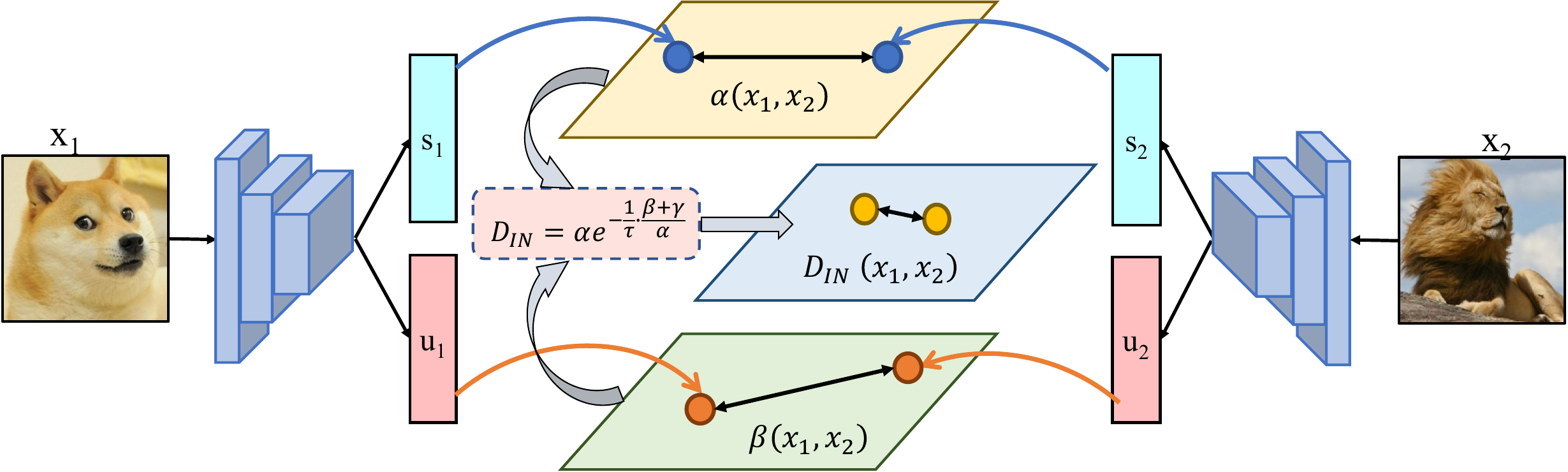}
\vspace{-2mm}
\caption{
An illustration of the proposed IDML framework. 
We employ a convolutional neural network to represent each image by a semantic embedding and an uncertainty embedding. 
We then use the distance between semantic embeddings as the semantic discrepancy and add the uncertainty embeddings for uncertainty measure.
The introspective similarity metric then uses the uncertainty level to weaken the semantic discrepancy to make a discreet similarity judgment.
} 
\label{fig:overall}
\figvspace
\end{figure*}

\subsection{Introspective Similarity Metric}
To facilitate an uncertainty-aware similarity metric, we first need to model the uncertainty in images.
Different from existing probabilistic deep embedding learning methods to model images as distributions, we propose to represent an image using a semantic embedding $\mathbf{s}$ and an uncertainty embedding $\mathbf{u}$, \emph{i.e.}, $\mathbf{y} = \{ \mathbf{s}, \mathbf{u} \}$.
The semantic embedding $\mathbf{s}$ describes the semantic characteristics of an image while the uncertainty embedding $\mathbf{u}$ models the ambiguity. 

For comparing two images $\mathbf{x}_1$ and $\mathbf{x}_2$, we define the semantic distance as $\alpha(\mathbf{x}_1, \mathbf{x}_2) = ||\mathbf{s}_1 - \mathbf{s}_2 ||_2$ similar to conventional DML methods but further compute a similarity uncertainty as $\beta(\mathbf{x}_1, \mathbf{x}_2) = ||\mathbf{u}_1 + \mathbf{u}_2 ||_2$.
Note that we add the vectors of the uncertainty embeddings before computing the norm instead of directly adding their norms.
The reason is that the uncertainty should depend on both concerning images.
For example, it might be difficult to differentiate an image from an elephant, but it can be affirmatively distinguished from a person.

When determining the semantic similarity between two images, an introspective metric needs to consider both the semantic distance and the similarity uncertainty. 
When not certain enough, the metric refuses to distinguish semantic discrepancies, as illustrated in Figure~\ref{fig:uncertain}. 
Formally, we consider it reckless to identify the semantic discrepancy between $\mathbf{x}_1$ and $\mathbf{x}_2$ when:
\begin{eqnarray}\label{equ:reckless}
\beta(\mathbf{x}_1, \mathbf{x}_2) + \gamma \geq \alpha(\mathbf{x}_1, \mathbf{x}_2),
\end{eqnarray}
where $\gamma > 0$ is the introspective bias indicating the introspective degree of the metric.
The positive value of $\gamma$ represents that the metric is still suspicious even if the image representation model provides no uncertainty.
We then define a strict introspective similarity metric as:
\begin{eqnarray}
\Tilde{D}_{IN}(\mathbf{x}_1,\mathbf{x}_2)= \alpha(\mathbf{x}_1,\mathbf{x}_2) \cdot I(\alpha(\mathbf{x}_1, \mathbf{x}_2) - \beta(\mathbf{x}_1, \mathbf{x}_2) - \gamma), 
\end{eqnarray}
where $I(x) $ is an indicator function which outputs 1 if $x>0$ and 0 otherwise.

However, the use of an indicator function is too strict and hard to optimize during training. 
We instead compare the semantic distance and the similarity uncertainty to define the relative uncertainty of two images:
\begin{eqnarray}
\text{r\_conf}(\mathbf{x}_1, \mathbf{x}_2) = \frac{\beta(\mathbf{x}_1, \mathbf{x}_2) + \gamma}{\alpha(\mathbf{x}_1, \mathbf{x}_2)}.
\end{eqnarray}
Note that the relative uncertainty is constantly positive.
We then use it to soften the semantic discrepancy to obtain our introspective similarity metric:
\begin{eqnarray}\label{equ:ecu}
{D}_{IN}(\mathbf{x}_1,\mathbf{x}_2)= \alpha(\mathbf{x}_1,\mathbf{x}_2) \cdot e^{(-\frac{1}{\tau} \  \text{r\_conf}(\mathbf{x}_1, \mathbf{x}_2))},
\end{eqnarray}
$\tau >0$ is a pre-defined hyperparameter to control the weakening degree. 
Note that the proposed introspective similarity metric does not satisfy the triangular equation and thus is not a mathematically strict metric.
We follow existing work~\cite{yuan2019signal,nguyen2010cosine} to still refer to it as a metric.

Intuitively, the proposed introspectively metric considers both the semantic distance and the similarity uncertainty between two images to conduct the final semantic discrepancies.
It generally produces a smaller distance for two images due to the awareness of the uncertainty. 
Given two pairs of images with the same semantic distance, the introspective metric distinguishes better for the pair with a smaller similarity uncertainty.
Also, when the uncertainty of two images outweighs the semantic distance to a great extent, the introspective metric simply outputs a near-zero semantic distance avoiding unnecessary influence on the network.

\subsection{Introspective Deep Metric Learning}
We demonstrate how to apply the proposed introspective metric to existing methods and present the overall framework of IDML, as illustrated in Figure~\ref{fig:overall}.

The semantic uncertainty naturally exists in images, yet it is hard to accurately describe the uncertainty for each image.
Moreover, many data augmentation methods~\cite{zhang2018mixup,yun2019cutmix,duan2018deep} in the literature expand the training data by generating synthetic images, which are known to possess multiple concepts.
Specifically, we employ the data mixing strategy~\cite{zhang2018mixup,yun2019cutmix} to demonstrate the advantage of our framework to deal with data with large uncertainty.

Although data uncertainty naturally exists in original images, we utilize the Mixup~\cite{zhang2018mixup} method to generate images with larger uncertainty to prove that our framework is capable of processing ambiguous data more effectively and achieving better performances. 
Specifically, we mix the original images $\mathbf{x}_1$ and $\mathbf{x}_2$ to obtain $\mathbf{x}_m=\lambda \cdot \mathbf{x}_1+(1-\lambda)\cdot \mathbf{x}_2$. However, different from the Mixup method~\cite{zhang2018mixup} which combines the labels of two images by $l_m=\lambda \cdot l_1+(1-\lambda)\cdot l_2$, we treat $l_m$ as a set which simultaneously includes $l_1$ and $l_2$, noted as $l_m=\{l_1,l_2\}$. On such condition, we define $l_i=l_j$ if $l_i \cap l_j \neq \emptyset$ and $l_i \neq l_j$ otherwise. We adopt the convolutional network to extract the feature embeddings of both original images and mixed images $\mathbf{y}_i=f(\mathbf{x}_i)=\{\mathbf{s}_i,\mathbf{u}_i\}$, where $\mathbf{s}_i$ and $\mathbf{u}_i$ denote the semantic feature embedding and the uncertainty feature embedding of $\mathbf{x}_i$, respectively.

The similarity computation of paired samples follows our introspective similarity metric.
In addition, our introspective similarity is compatible with a variety of loss formulations. For example, we can employ the margin loss with the distance-weighted sampling strategy~\cite{wu2017sampling} to optimize the pairwise distances:
\begin{eqnarray}
&& J_{m}(\mathbf{y},\mathbf{L})\! = \!\!\sum_{l_i = l_j} [{D}_{IN} (\mathbf{x}_i, \mathbf{x}_j)\!-\!\xi]_+ \nonumber \\
&& \!\!-\!\! \sum_{l_i \neq l_j} I(p({D}_{IN} (\mathbf{x}_i, \mathbf{x}_j))) [\omega \!-\! {D}_{IN} (\mathbf{x}_i, \mathbf{x}_j)]_+,
\end{eqnarray}
where ${D}_{IN} (\mathbf{x}_i, \mathbf{x}_j)$ follows \eqref{equ:ecu}, the random variable $I(p)$ has a probability of $p$ to be 1 and 0 otherwise, $p(d)=\min(\phi,d^{2-n}[1-\frac{1}{4}d^2]^{\frac{3-n}{2}})$, $[\cdot]_+ = \max(\cdot,0)$, $\xi$ and $\omega$ are two pre-defined margins, and $\phi$ is a positive constant.

In addition, proxy-based losses such as the softmax loss~\cite{deng2019arcface,sun2020circle} fits our similarity metric as well. It should be noted that different from the Euclidean distance, various proxy-based losses compute the cosine similarity $C(\mathbf{x}_i, \mathbf{p}_j)$ between an image $\mathbf{x}_i$ and a class-level representative $\mathbf{p}_j$. Therefore, we conduct the confidence decay of the semantic similarity in another way:
\begin{eqnarray}
{C}_{IN}(\mathbf{x}_i, \mathbf{p}_j)=1  -  (1-C(\mathbf{x}_i, \mathbf{p}_j))\cdot e^{(-\frac{1}{\tau} \  \text{r\_conf}(\mathbf{x}_i, \mathbf{p}_j))},
\end{eqnarray}
where $\text{r\_conf}(\mathbf{x}_i, \mathbf{p}_j)$ denotes the relative uncertainty between the image and the representative, and we provide the softmax loss based on our similarity metric:
\begin{eqnarray}
J_{s}(\mathbf{y},\mathbf{L}) = \frac{1}{N}\sum_{i=1}^{N}(-log\frac{\sum_{l_{\mathbf{p}_j}=l_{i}} e^{C_{IN}(\mathbf{x}_i,\mathbf{p}_j)}}{\sum_{l_{\mathbf{p}_j}\neq l_{i}}e^{C_{IN}(\mathbf{x}_i,\mathbf{p}_j)}}),
\end{eqnarray}
where $l_{\mathbf{p}_j}$ is the category of the class representative $\mathbf{p}_j$.

For inference, we directly use the Euclidean distance between semantic embeddings as the similarity metric and can optionally use the uncertainty embedding to indicate the uncertainty level.

\newcommand{\tablesize}{\small}
\newcommand{\arraysep}{\renewcommand\arraystretch{1.3}}

 \begin{table*}[!t] \tablesize
 \centering
 \caption{Experimental results (\%) on the CUB-200-2011, Cars196, and Stanford Online Products datasets compared with state-of-the-art methods.  * denotes our reproduced results under the same settings.}
 \label{tab:sota}
 \vspace{-7pt}
 \setlength\tabcolsep{3.5pt}
 \arraysep
 \begin{tabular}{lc|ccccc|ccccc|ccccc}
 \hline
  & & \multicolumn{5}{|c}{CUB-200-2011} & \multicolumn{5}{|c}{Cars196} & \multicolumn{5}{|c}{Stanford Online Products} \\
  \hline
 Method & Setting & R@1 & R@2 & NMI & RP &M@R & R@1 & R@2 & NMI & RP & M@R & R@1 & R@10 & NMI & RP & M@R \\
 \hline
 ABE-8~\cite{kim2018attention} & 512G & 60.6 & 71.5 & - & - & - & 85.2 & 90.5 & - & - & - & 76.3 & 88.4 & - & - & -\\
 Ranked~\cite{wang2019ranked} & 1536BN & 61.3 & 72.7 & 66.1 & - & - & 82.1 & 89.3 & 71.8 & - & - & 79.8 & 91.3 & 90.4 & - & -\\
 DREML~\cite{xuan2018deep} & 9216R & 63.9 & 75.0 & 67.8 & - & - & 86.0 & 91.7 & 76.4 & - & - & - & - & - & - & -\\
 SoftTriple~\cite{qian2019softtriple} & 512BN & 65.4 & 76.4 & 69.3 & - & - & 84.5 & 90.7 & 70.1 & - & - & 78.3 & 90.3 & 92.0 & - & -\\
 D \& C~\cite{sanakoyeu2019divide} & 128R & 65.9 & 76.6 & 69.6 & - & - & 84.6 & 90.7 & 70.3 & - & - & 75.9 & 88.4 & 90.2 & - & -\\
 MIC~\cite{roth2019mic} & 128R & 66.1 & 76.8 & 69.7 & - & - & 82.6 & 89.1  & 68.4 & - & - & 77.2 & 89.4 & 90.0 & - & -\\
 RankMI~\cite{kemertas2020rankmi} & 128R & 66.7 & 77.2 & 71.3 & - & - & 83.3 & 89.8 & 69.4 & - & - & 74.3 & 87.9 & 90.5 & - & -\\
 CircleLoss~\cite{sun2020circle} & 512R & 66.7 & 77.4 & - & - & - & 83.4 & 89.8 & - & - & - & 78.3 & 90.5 & - & - & -\\
 PADS~\cite{roth2020pads} & 128BN & 67.3 & 78.0 & 69.9 & - & - & 83.5 & 89.7 & 68.8 & - & - & 76.5 & 89.0 & 89.9 & - & -\\
 DIML~\cite{zhao2021towards} & 512R & 68.2 & - & - & 37.9 & 26.9 & 87.0 & - & - & 39.0 & 29.4 & 79.3 & - & - & 46.4 & 43.2\\
 DCML~\cite{zheng2021deep} & 512R & 68.4 & 77.9 & 71.8 & - & - & 85.2 & 91.8 & 73.9 & - & - & 79.8 & 90.8 & 90.8 & - & -\\
 DRML~\cite{zheng2021deepr} & 512BN & 68.7 & 78.6 & 69.3 & - & - & 86.9 & 92.1 & 72.1 & - & - & 79.9 & 90.7 & 90.1 & - & -\\
 ProxyNCA++~\cite{teh2020proxynca++}  & 512R & 69.0 & 79.8 & \color{red}73.9 & - & - & 86.5 & 92.5 & 73.8 & - & - & \color{blue}80.7 & \color{blue}92.0 & - & - & -\\ 
 DiVA~\cite{milbich2020diva} & 512R & 69.2 & 79.3 & 71.4 & - & - & 87.6 & 92.9 & 72.2 & - & - & 79.6 & 91.2 & 90.6 & - & - \\
 NIR~\cite{roth2022non} & 512R & \color{blue}{70.5} & \color{red}{80.6} & 72.5 & - & - & \color{blue}89.1 & \color{blue}93.4 & 75.0 & - & - & \color{blue}80.7 & 91.5 & 90.9 & - & - \\
 \hline
 Triplet-SH*~\cite{schroff2015facenet} & 512R & 63.6 & 75.5 & 67.9 & 35.1 & 24.0 & 70.8 & 81.7 & 64.8 & 31.7 & 21.1 & 76.5 & 89.1 & 89.7 & 51.3 & 48.4\\
 IDML-TSH & 512R & \textbf{65.3} & \textbf{76.5} & \textbf{69.5} & \textbf{36.2} & \textbf{25.0} & \textbf{73.7} & \textbf{84.0} & \textbf{67.3} & \textbf{33.8} & \textbf{24.1} & \textbf{77.4} & \textbf{89.4} & \textbf{90.1} & \textbf{51.9} & \textbf{49.0}\\
 \hline
 ProxyNCA*~\cite{movshovitz2017no} & 512R & 64.6 & 75.6 & 69.1 & 35.5 & 24.7 & 82.6 & 89.0 & 66.4 & 33.5 & 23.5 & 77.0 & 89.1 & 89.5 & 51.9 & 49.0\\
 IDML-PN & 512R & \textbf{66.0} & \textbf{76.4} & \textbf{70.1} & \textbf{36.5} & \textbf{25.4} & \textbf{85.5} & \textbf{91.3} & \textbf{69.0} & \textbf{36.1} & \textbf{26.4} & \textbf{78.3} & \textbf{90.1} & \textbf{89.9} & \textbf{53.0} & \textbf{49.9}\\
 \hline
 FastAP*~\cite{cakir2019deep} & 512R & 65.1 & 75.4 & 68.5 & 35.9 & 24.1 & 81.6 & 88.5 & 68.8 & 35.1 & 25.2 & 75.9 & 89.2 & 89.7 & 50.1 & 46.8\\
 IDML-FAP & 512R & \textbf{66.4} & \textbf{76.4} & \textbf{69.7} & \textbf{36.7} & \textbf{25.5} & \textbf{83.9} & \textbf{89.9} & \textbf{71.9} & \textbf{36.5} & \textbf{26.7} & \textbf{76.8} & \textbf{89.7} & \textbf{90.9} & \textbf{50.9} & \textbf{47.9}\\
 \hline
 Contrastive*~\cite{hu2014discriminative} & 512R & 65.6 & 76.5 & 68.9 & 36.5 & 24.7 & 82.7 & 89.6 & 69.5 & 35.8 & 25.7 & 76.4 & 88.5 & 88.9 & 50.9 & 47.9 \\
 IDML-Con & 512R & \textbf{67.2} & \textbf{77.6} & \textbf{71.3} & \textbf{37.5} & \textbf{25.7} & \textbf{85.5} & \textbf{91.5} & \textbf{72.5} & \textbf{38.8} & \textbf{29.0} & \textbf{77.3} & \textbf{89.7} & \textbf{90.0} & \textbf{51.7} & \textbf{48.5} \\
 \hline
 Margin-DW*~\cite{wu2017sampling} & 512R & 65.9 & 77.0 & 69.5 & 36.0 & 24.9 & 82.6 & 88.7 & 69.3 & 36.4 & 26.5 & 78.5 & 89.9 & 90.1 & 53.4 & 50.2\\
 IDML-MDW & 512R & \textbf{67.9} & \textbf{78.3} & \textbf{72.1} & \textbf{37.2} & \textbf{26.1} & \textbf{86.1} & \textbf{91.7} & \textbf{73.0} & \textbf{39.2} & \textbf{29.7} & \textbf{79.4} & \textbf{90.6} & \textbf{91.0} & \color{blue}\textbf{53.7} & \textbf{50.4}\\
 \hline
 Multi-Sim*~\cite{wang2019multi} & 512R & 67.3 & 78.2 & 72.7 & 36.6 & 25.5 & 83.3 & 90.9 & 72.2 & 37.4 & 27.4 & 78.1 & 90.0 & 89.9 & 52.9 & 49.9\\
 IDML-MS & 512R & \textbf{69.0} & \textbf{79.5} & \textbf{73.5} & \color{blue}\textbf{38.5} & \textbf{27.2} & \textbf{86.3} & \textbf{92.2} & \textbf{74.1} & \textbf{40.0} & \textbf{30.8} & \textbf{79.7} & \textbf{91.4} & \color{blue}\textbf{91.2} & \color{blue}\textbf{53.7} & \color{blue}\textbf{50.9}\\
 \hline
 ProxyAnchor*~\cite{kim2020proxy} & 512R & 69.0 & 79.4 & 72.3 & \color{blue}38.5 & \color{blue}27.5 & 87.3 & 92.7 & \color{blue}75.7 & \color{blue}40.9 & \color{blue}31.8 & 79.5 & 91.1 & 91.0 & \color{blue}53.7 & 50.5\\
 IDML-PA & 512R & \color{red}\textbf{70.7} & \color{blue}\textbf{80.2} & \color{blue}\textbf{73.5} & \color{red}\textbf{39.3} & \color{red}\textbf{28.4} & \color{red}\textbf{90.6} & \color{red}\textbf{94.5} & \color{red}\textbf{76.9} & \color{red}\textbf{42.6} & \color{red}\textbf{33.8} & \color{red}\textbf{81.5} & \color{red}\textbf{92.7} & \color{red}\textbf{92.3} & \color{red}\textbf{54.8} & \color{red}\textbf{51.3}\\
 \hline
 \end{tabular}
 \tablevspace
 \end{table*}

\subsection{Gradient Analysis}
We provide a gradient analysis to demonstrate the effect of our introspective similarity metric on the learning of semantic embeddings.
Intuitively, our proposed similarity metric results in reduced semantic discrepancies to impose fewer influences for uncertain samples. 
Therefore, we present a detailed gradient analysis of the parameters of the semantic embedding layer $W^t$ with the Euclidean-distance-based methods for example. 
Firstly, we formulate the backpropagation process in the $t$-th iteration as follows:
\begin{eqnarray}
\frac{\partial J}{\partial W^{t}}= \frac{\partial J}{\partial \mathbf{S}}\cdot \frac{\partial \mathbf{S}}{\partial W^t}= \frac{\partial J}{\partial {D}_{IN}}\cdot \frac{\partial {D}_{IN}}{\partial \mathbf{S}}\cdot \frac{\partial \mathbf{S}}{\partial W^t},
\end{eqnarray}
where $J$ and $\mathbf{S}$ denote the loss function and the semantic embeddings for simplicity. Actually, the partial term $\frac{\partial J}{\partial D_{IN}}$ depends on the form of the loss function while $\frac{\partial \mathbf{S}}{\partial W^t}$ is merely relevant to the architecture of the network. On such condition, $\frac{\partial {D}_{IN}}{\partial \mathbf{S}}$ contributes to the update level of the parameters, and a larger gradient undoubtedly represents a larger update. Next, we formulate the above gradient based on a paired samples $\{\mathbf{y}_1=(\mathbf{s}_1,\mathbf{u_1}),\mathbf{y}_2=(\mathbf{s}_2,\mathbf{u_2})\}$, which is as follows: (${D}_{IN}$ is the same as \eqref{equ:ecu} and we set $\tau=1$ and $\gamma=0$ for simplicity.)
\begin{eqnarray}
&&\!\!\!\!\!\!\!\!\!\!H(\mathbf{u}_1,\mathbf{u}_2)\!=\!\frac{\partial {D}_{IN}(\mathbf{x}_1,\mathbf{x}_2)}{\partial \mathbf{s}_1} \nonumber \\
&&\!\!\!\!\! \!\!\!\!\! \!=\!\frac{\partial ||\mathbf{s}_1-\mathbf{s}_2||_2}{\partial \mathbf{s}_1}\! \cdot\! e^{-\frac{||\mathbf{u}_1+\mathbf{u}_2||_2}{||\mathbf{s}_1-\mathbf{s}_2||_2}}\!\cdot\!
(1+\frac{||\mathbf{u}_1+\mathbf{u}_2||_2}{||\mathbf{s}_1-\mathbf{s}_2||_2})
\end{eqnarray}
where $H(\mathbf{u}_1,\mathbf{u}_2)$ is a function of $\{\mathbf{u}_1,\mathbf{u}_2\}$ if the semantic feature embeddings $\{\mathbf{s}_1,\mathbf{s}_2\}$ are fixed, $\frac{\partial ||\mathbf{s}_1-\mathbf{s}_2||_2}{\partial \mathbf{s}_1}$ denotes the gradient for the original Euclidean distance form, and $e^{-\frac{||\mathbf{u}_1+\mathbf{u}_2||_2}{||\mathbf{s}_1-\mathbf{s}_2||_2}}\cdot(1+\frac{||\mathbf{u}_1+\mathbf{u}_2||_2}{||\mathbf{s}_1-\mathbf{s}_2||_2})$ connects with the commonly used monotone decreasing function $g(x)=e^{-x}\cdot(1+x)$ when $x>0$, which remains decreasing when $\mathbf{u}_1>0$.

Therefore, $H(\mathbf{u}_1,\mathbf{u}_2)$ decreases when the length of the uncertainty embedding $\mathbf{u}_1$ increases. It indicates that images with more uncertainty result in smaller gradients, thus imposing fewer influences on the network. Specially, when $||\mathbf{u}_1+\mathbf{u}_2||_2=0$, denoting the absolute certainty of samples, $H(\mathbf{u}_1,\mathbf{u}_2)=\frac{\partial ||\mathbf{s}_1-\mathbf{s}_2||_2}{\partial \mathbf{s}_1}$, turning back to the form of the original Euclidean gradient.
 It denotes that our similarity metric maintains the original judgment for certain samples.

\begin{table*}[t] \tablesize
\centering
\caption{Analysis of different components of IDML on the CUB-200-2011, Cars196, and Stanford Online Products datasets.}
\label{tab:component}
\vspace{-7pt}
\setlength\tabcolsep{8.5pt}
\arraysep
\begin{tabular}{l|cccc|cccc|cccc}
\hline
& \multicolumn{4}{|c}{CUB-200-2011} & \multicolumn{4}{|c}{Cars196} & \multicolumn{4}{|c}{Stanford Online Products}\\
 \hline
Method & R@1 & NMI & RP & M@R & R@1 & NMI & RP & M@R & R@1 & NMI & RP & M@R\\
\hline
Margin-DW~\cite{wu2017sampling} & 65.9 & 69.5 & 36.0 & 24.9  & 82.6 & 69.3 & 36.4 & 26.5 & 78.5 & 90.1 & 53.4 & 50.2\\ 
Mixup-MDW & 67.1 & 71.6 & 36.7 & 25.5 & 84.7 & 72.4 & 38.0 & 28.0 & 79.1 & 90.5 & 53.6 & \textbf{50.4}\\
ISM-MDW & 67.0 & 71.4 & 36.9 & 25.7 & 84.4 & 71.9 & 37.9 & 28.1 & 78.9 & 90.4 & 53.6 & 50.3\\
PEL-MDW~\cite{oh2018modeling} & 63.3 & 67.1 & 34.6 & 24.2 & 80.4 & 67.1 & 34.8 & 25.5 & 76.4 & 88.7 & 51.2 & 48.7\\
PEL-Mixup-MDW & 64.5 & 68.6 & 35.3 & 24.7 & 82.3 & 68.9 & 35.9 & 26.1 & 77.2 & 89.4 & 52.1 & 49.3\\
IDML-MDW & \textbf{67.9} & \textbf{72.1} & \textbf{37.2} & \textbf{26.1} & \textbf{86.1} & \textbf{73.0} & \textbf{39.2} & \textbf{29.7} & \textbf{79.4} & \textbf{91.0} & \textbf{53.7} & \textbf{50.4}\\
\hline
ProxyAnchor~\cite{kim2020proxy} & 69.0 & 72.3 & 38.5 & 27.5 & 87.3 & 75.7 & 40.9 & 31.8 & 79.5 & 91.0 & 53.7 & 50.5\\
Mixup-PA & 69.8 & 73.0 & 39.1 & 28.1 & 88.5 & 75.8 & 41.0 & 32.1 & 80.6 & 91.8 & 54.4 & 50.7\\
ISM-PA & 69.5 & 73.1 & 38.9 & 28.0 & 88.8 & 75.8 & 41.2 & 32.2 & 80.3 & 91.8 & 54.3 & 50.9\\
PEL-PA~\cite{oh2018modeling} & 64.9 & 67.1 & 34.5 & 23.7 & 83.4 & 66.4 & 34.4 & 24.9 & 76.8 & 89.7 & 51.8 & 48.7 \\
PEL-Mixup-PA  & 65.7 & 68.0 & 35.6 & 24.7 & 84.5 & 66.6 & 34.6 & 25.1 & 77.9 & 90.5 & 52.6 & 49.9 \\
IDML-PA & \textbf{70.7} & \textbf{73.5} & \textbf{39.3} & \textbf{28.4} & \textbf{90.6} & \textbf{76.9} & \textbf{42.6} & \textbf{33.8} & \textbf{81.5} & \textbf{92.3} & \textbf{54.8} & \textbf{51.3}\\
\hline
\end{tabular}
\vspace{-3mm}
\end{table*}

\section{Experiments}

In this section, we conducted various experiments to evaluate the performance of our IDML framework on image retrieval.
We show that employing the proposed introspective similarity metric consistently improves the performance of existing deep metric learning and data mixing methods.
We also provide in-depth analyses of the effectiveness of our framework.

\subsection{Settings}
We first evaluated our framework under the conventional deep metric learning setting~\cite{song2016deep,wang2019multi} and conducted experiments on three widely-used datasets: CUB-200-2011~\cite{wah2011caltech}, Cars196~\cite{krause20133d}, and Stanford Online Products~\cite{song2016deep}. 
We provide the detailed dataset splits in the supplementary material.
We adopted the ImageNet~\cite{russakovsky2015imagenet} pretrained ResNet-50~\cite{he2016deep} as the backbone and two randomly initialized fully connected layers to obtain the semantic embedding and uncertainty embedding, respectively.
We set the embedding size to 512 for the main experiments.
The training images were first resized to $256\times256$ and then augmented with random cropping to $224\times224$ as well as random horizontal flipping with the probability of $50\%$. 
We employed Mixup~\cite{zhang2018mixup} for our framework to generate images with large uncertainty for training.
We fixed the batch size to $120$ and used the AdamW optimizer with the learning rate of $10^{-5}$. 
We set $\gamma=3$ for Cars196 and $\gamma=0$ for the other datasets and fixed $\epsilon=5$ for all the datasets during training. 
We adopted the original similarity metric without our uncertainty-aware term for testing. 
The reported evaluation metrics include Recall@Ks, normalized mutual information (NMI), R-Precision (RP), and Mean Average Precision at R (M@R).
See Musgrave~\emph{et al.}~\cite{musgrave2020metric} for more details.

\subsection{Dataset}
For image retrieval and clustering, we followed existing DML methods~\cite{song2016deep, zheng2019hardness, wang2019multi, kim2020proxy} to conduct experiments on the CUB-200-2011~\cite{wah2011caltech}, Cars196~\cite{krause20133d}, and Stanford Online Products~\cite{song2016deep}. 
CUB-200-2011 contains 11,788 images of 200 bird species. 
We used the first 100 species with 5,864 images for training and the rest 100 species with 5,924 images are for testing. 
Cars196 includes 16,183 images of 196 car models. 
We used the first 96 classes with 8,054 images for training and the rest 98 classes with 8,131 images for testing. 
Stanford Online Products is relatively large and contains 120,053 images of 22,634 products. 
We used the first 11,318 products with 59,551 images in the training set and the rest 11,318 products with 60,502 images for testing.

\begin{table*}[t] \tablesize 
  \centering
  \caption{Experimental results with various forms of uncertain data.}
  \label{tab:uncertainty} 
\vspace{-7pt}
\setlength\tabcolsep{8.3pt}
\arraysep
    \begin{tabular}{l|cccc|cccc|cccc}
    \hline
    \multicolumn{1}{c|}{\multirow{2}[0]{*}{Method}} & \multicolumn{4}{c|}{CUB-200-2011} & \multicolumn{4}{c|}{Cars196} & \multicolumn{4}{c}{Stanford Online Products}  \\ \cline{2-13}
          & R@1   & NMI   & RP    & M@R   & R@1   & NMI   & RP    & M@R  & R@1   & NMI   & RP    & M@R  \\
    \hline
    Baseline & 69.0  & 72.3  & 38.5  & 27.5  & 87.3  & 75.7  & 40.9  & 31.8  & 79.5  & 91.0  & 53.7  & 50.5 \\
    Baseline + ISM & 69.5  & 73.1  & 38.9  & 28.0  & 88.8  & 75.8  & 41.2  & 32.2  & 80.3  & 91.8  & 54.3  & 50.9 \\
    Low-res (200$\times$200) & 67.4  & 71.3  & 37.7  & 26.2  & 87.0  & 75.3  & 40.8  & 31.2  & 78.7  & 90.0  & 52.6  & 49.1 \\
    Low-res + ISM & 68.9  & 71.9  & 38.2  & 27.4  & 89.0  & 76.2  & 41.3  & 32.5  & 79.3  & 90.9  & 53.4  & 50.4 \\
    Blur (p=0.5)  & 69.0  & 72.4  & 38.5  & 27.7  & 88.2  & 75.8  & 41.1  & 32.0  & 79.7  & 91.0  & 53.9  & 50.5 \\
    Blur + ISM & 69.2  & 72.5  & 38.6  & 27.7  & 89.6  & 76.4  & 41.8  & 32.8  & 79.7  & 91.1  & 53.9  & 50.3 \\
    Occlusion (p=0.5) & 69.3  & 72.4  & 38.6  & 27.9  & 87.9  & 75.7  & 41.1  & 31.8  & 80.2  & 91.2  & 54.2  & 50.7 \\
    Occlusion + ISM & 69.6  & 72.8  & 38.8  & 28.0  & 89.2  & 76.4  & 41.5  & 32.6  & 80.6  & 91.7  & 54.6  & 50.9 \\
    Mixup & 69.8  & 73.0  & 39.1  & 28.1  & 88.5  & 75.8  & 41.0  & 32.1  & 80.6  & 91.8  & 54.4  & 50.7 \\
    Mixup + ISM & \textbf{70.7} & \textbf{73.5} & \textbf{39.3} & \textbf{28.4} & \textbf{90.6} & \textbf{76.9} & \textbf{42.6} & \textbf{33.8} & \textbf{81.5} & \textbf{92.3} & \textbf{54.8} & \textbf{51.3} \\
    \hline
    \end{tabular}%
   \vspace{-5mm}
\end{table*}%

\newcommand\figwidth{0.323}

\begin{figure*}[t]
\centering
\subcaptionbox{ CUB-200-2011\label{subfig:cub}}{
\includegraphics[width=\figwidth\textwidth]{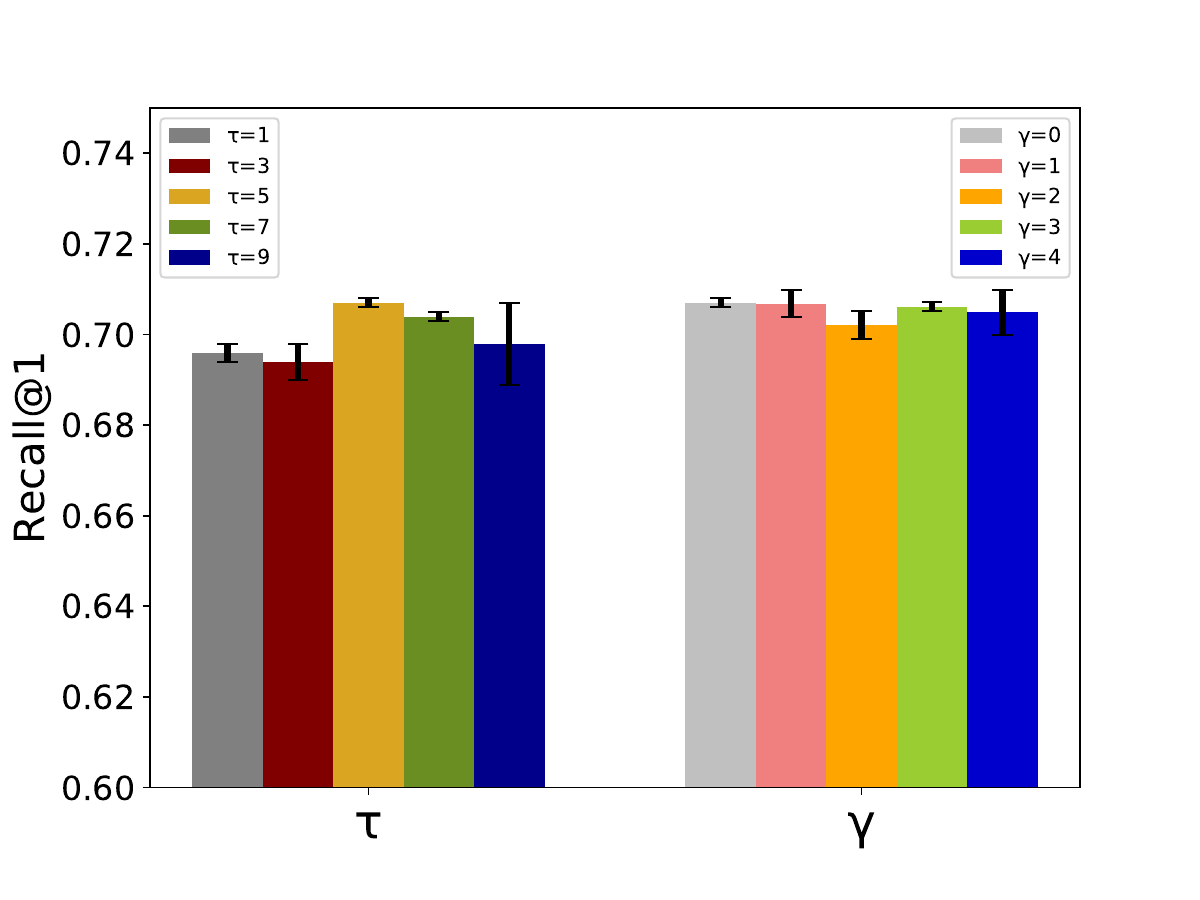}
}\hfill
\subcaptionbox{ Cars196\label{subfig:car}}{
\includegraphics[width=\figwidth\textwidth]{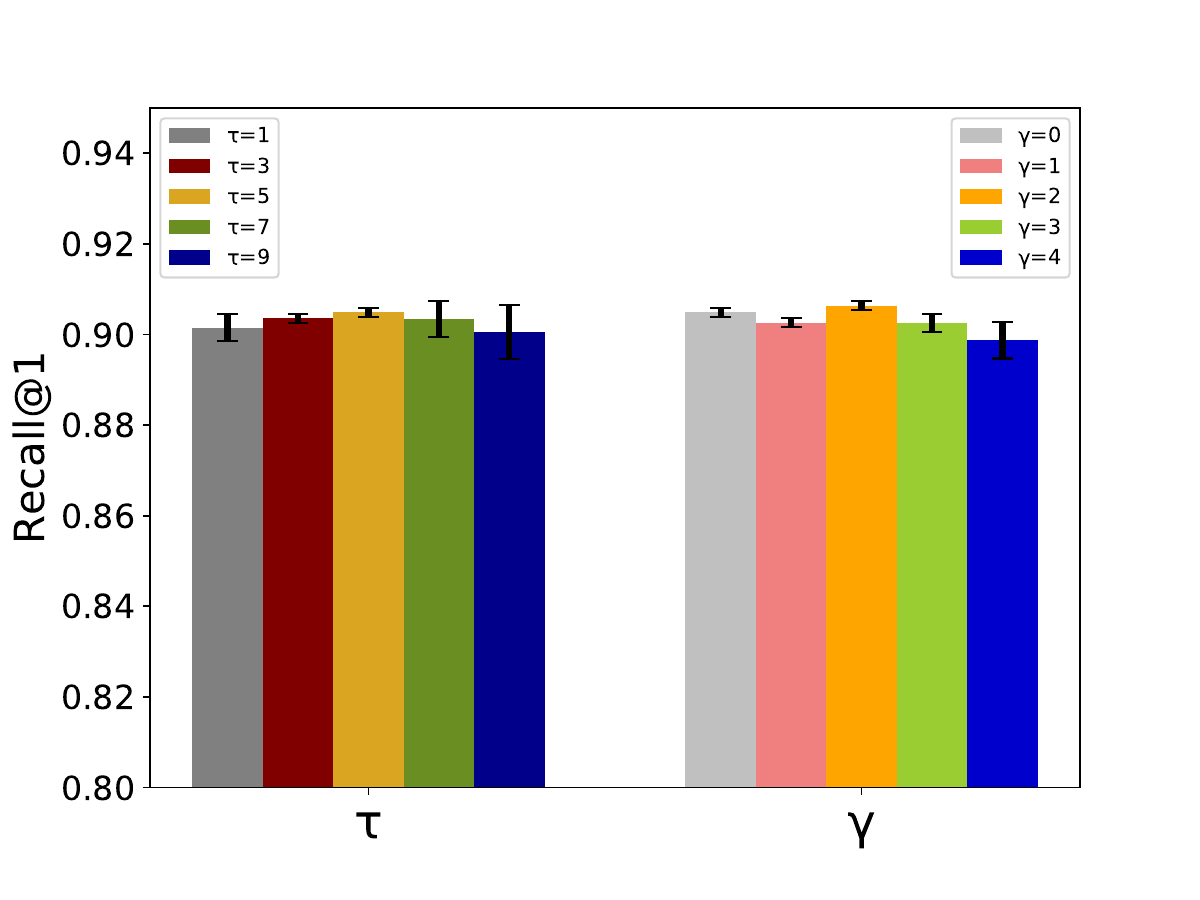}
}\hfill
\subcaptionbox{ 
Stanford Online Products\label{subfig:sop}}{
\includegraphics[width=\figwidth\textwidth]{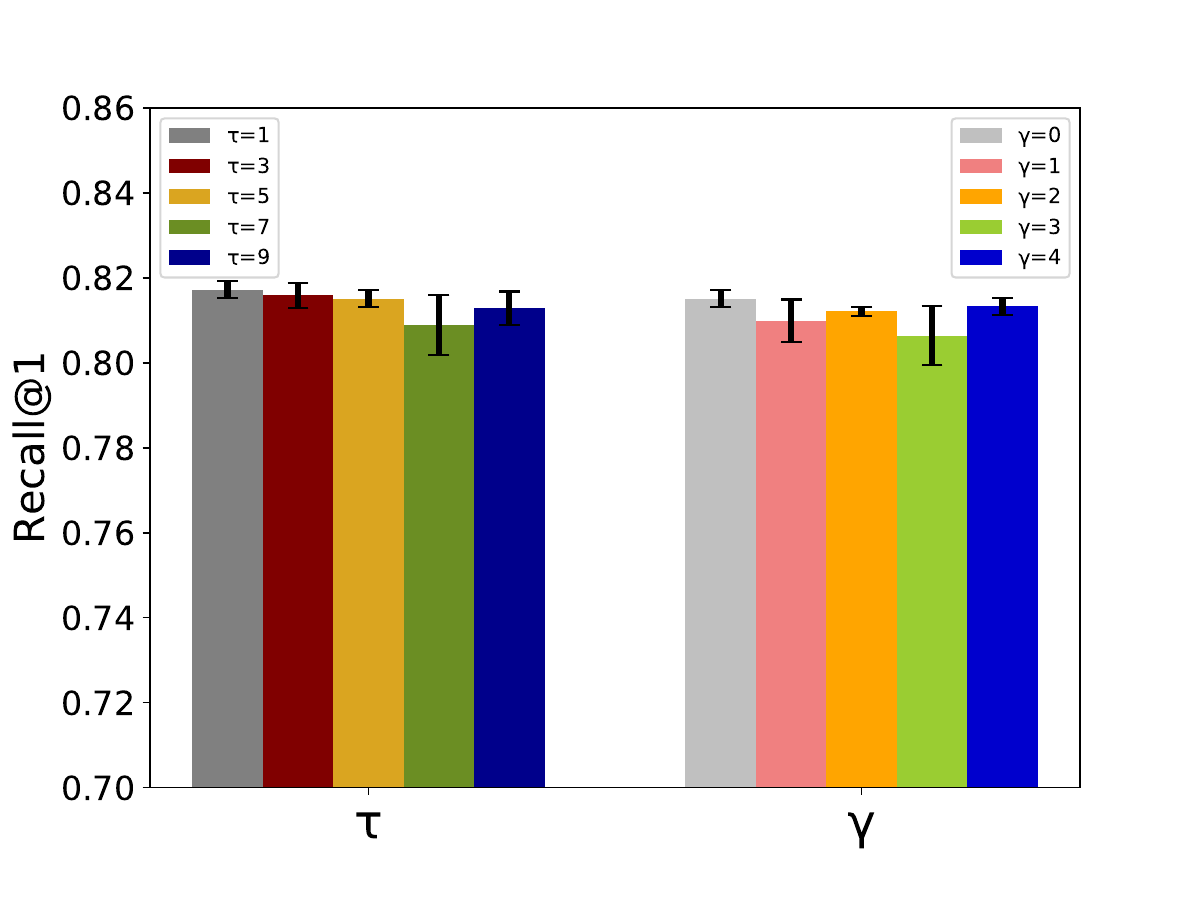}
}
\vspace{-3mm}
\caption{Impact of metric parameters on the CUB-200-2011, Cars196 and Stanford Online Products datasets.}
\label{fig:para}
\figvspace
\end{figure*}

\subsection{Main Results}
We evaluated the proposed IDML framework under the conventional deep metric learning setting and compared it with state-of-the-art methods.
To demonstrate the versatility of our framework, we applied the introspective similarity metric to various loss functions, including the triplet loss with the semi-hard sampling (Triplet-SH)~\cite{schroff2015facenet}, the ProxyNCA loss~\cite{movshovitz2017no}, the FastAP loss~\cite{cakir2019deep}, the contrastive loss~\cite{hu2014discriminative}, the margin loss with the distance-weighed sampling (Margin-DW)~\cite{wu2017sampling}, the multi-similarity loss (Multi-Sim)~\cite{wang2019multi}, and the ProxyAnchor loss~\cite{kim2020proxy}. 

Table~\ref{tab:sota} shows the experimental results on the CUB-200-2011~\cite{wah2011caltech}, Cars196~\cite{krause20133d}, and Stanford Online Products~\cite{song2016deep} datasets. 
The n-BN/R denotes the model setting where n is the embedding size and BN, R represents BN-Incertion~\cite{ioffe2015batch} and ResNet-50~\cite{he2016deep}, respectively. 
The bold numbers highlight the improvement of our framework compared with the original method, and red numbers indicate the best results.
We observe that our framework achieves a constant performance boost to all the associated methods. 
Furthermore, we attain state-of-the-art performance on all three datasets by applying our framework to the ProxyAnchor loss, which surpasses the original performance by $3.3\%$ at Recall@1 and $2.0\%$ at M@R on the Cars196 dataset, respectively.
This is because the proposed similarity metric is aware of the data uncertainty in images so that the uncertain samples only provide limited training signals.

\subsection{Analysis} \label{analysis}

\textbf{Ablation Study of Different Components:}
We conducted experiments with the margin loss and ProxyAnchor loss to analyze the effect of different components of our framework, as shown in Table~\ref{tab:component}.
We first applied Mixup to the baseline method (Mixup-MDW) without using our introspective similarity metric and then only employed the proposed metric without mixup (ISM-MDW).
We see that the Mixup method and the proposed ISM can independently boost the performances of the baseline method.
Our IDML framework further improves the performance by combining Mixup and our ISM. 
Furthermore, we reproduced the probabilistic embedding learning (PEL) framework~\cite{oh2018modeling} on each loss (PEL-MDW) and also equipped it with Mixup (PEL-Mixup-MDW) for fair comparisons with our framework.
We observe that it achieves lower performance than the baseline method, and further using Mixup improves the performance.
The performance drop might result from the compromise of discriminativeness when representing images as distributions.
Differently, IDML uses an uncertainty embedding to model the uncertainty which does not affect the discriminativeness of the semantic embedding.

\textbf{Uncertainty of Other Forms:}
In addition to Mixup, we further conducted experiments when training with lowered-resolution, blurred, and occluded images, as shown in Table~\ref{tab:uncertainty}.
Though certain augmentations (\emph{e.g.}, low-resolution) reduce the performance of the baseline, further applying our ISM consistently attains better results than training without these augmentations.
For example, adopting the low-resolution augmentation achieves 67.4$\%$ at Recall@1 on the CUB-200-211 dataset, which is 1.6$\%$ lower than the baseline, while further applying ISM improves the performance by $1.5\%$ and is close to the original baseline method.
This verifies the effectiveness of our method to deal with various forms of uncertain data.

\begin{table}[t] \small
\centering
\caption{Effect of different metric formulations during training.}
\label{tab:dissimilar}
\vspace{-7pt}
\setlength\tabcolsep{5.2pt}
\arraysep
\begin{tabular}{l|l|cccc}
\hline
Dataset & Training Metric & R@1 & NMI & RP & M@R\\
\hline
\multirow{3}{*}{CUB-200-2011} & Euclidean & 69.0 & 72.3 & 38.5 & 27.5\\ 
  & ISM-Dis~\eqref{eq:dis} & 69.2 & 72.0 & 38.7 & 28.1\\
  & ISM-Sim~\eqref{eq:sim} & \textbf{70.7} & \textbf{73.5} & \textbf{39.3} & \textbf{28.4}\\
\hline
\multirow{3}{*}{Cars196} & Euclidean & 87.3 & 75.7 & 40.9 & 31.8\\
  & ISM-Dis~\eqref{eq:dis} & 89.4 & 75.4 & 41.6 & 32.5\\
  & ISM-Sim~\eqref{eq:sim} & \textbf{90.6} & \textbf{76.9} & \textbf{42.6} & \textbf{33.8}\\
\hline
\multirow{3}{*}{SOP} & Euclidean & 79.5 & 91.0 & 53.7 & 50.5\\
  & ISM-Dis~\eqref{eq:dis} & 80.2 & 91.3 & 54.0 & 50.6\\
  & ISM-Sim~\eqref{eq:sim} & \textbf{81.5} & \textbf{92.3} & \textbf{54.8} & \textbf{51.3}\\
\hline
\end{tabular}
\tablevspace
\end{table}

\textbf{Effect of the Hyper-parameters $\gamma$ and $\tau$:}
$\gamma$ determines the introspective bias and $\tau$ controls the weakening degree in our introspective similarity metric. 
They jointly affect the final performance of our framework. 
We experimentally evaluated their impacts as demonstrated in Figure~\ref{fig:para}. 
We first fixed $\gamma$ to $0$ and set $\tau$ to 1, 3, 5, 7, 9. 
We see that our framework achieves the best recall@1 when $\tau=5$ for both datasets, indicating the favor of a modest weakening degree.
In addition, we fixed $\tau=5$ and set $\gamma$ to 0, 1, 2, 3, 4 for training. 
The experimental results vary on the two datasets.
Specifically, our framework achieves the best performance when $\gamma=0$ on the CUB-200-2011 dataset while $\gamma=3$ on the Cars196 dataset.
This indicates that the metric is more discreet when comparing images on the Cars196 dataset.

\begin{table}[t] \small
\centering
\caption{Effect of different metric formulations during testing.}
\label{tab:ISM}
\vspace{-7pt}
\setlength\tabcolsep{3.6pt}
\arraysep
\begin{tabular}{l|l|cccc}
\hline
Dataset & Testing Metric & R@1 & NMI & RP & M@R\\
\hline
\multirow{3}{*}{CUB-200-2011} & Euclidean (baseline) & 69.0 & 72.3 & 38.5 & 27.5\\ 
  & ISM (IDML) & 69.8 & 73.1 & 39.0 & 27.8 \\ 
  & Euclidean (IDML) & \textbf{70.7} & \textbf{73.5} & \textbf{39.3} & \textbf{28.4}\\
\hline
\multirow{3}{*}{Cars196} & Euclidean (baseline) & 87.3 & 75.7 & 40.9 & 31.8\\
  & ISM (IDML) & 89.9 & 76.2 & 42.3 & 33.3\\
  & Euclidean (IDML) & \textbf{90.6} & \textbf{76.9} & \textbf{42.6} & \textbf{33.8}\\
\hline
\multirow{3}{*}{SOP} & Euclidean (baseline) & 79.5 & 91.0 & 53.7 & 50.5\\
  & ISM (IDML) & 80.7 & 92.1 & 54.2 & 50.8\\
  & Euclidean (IDML) & \textbf{81.5} & \textbf{92.3} & \textbf{54.8} & \textbf{51.3}\\
\hline
\end{tabular}
\tablevspace
\end{table}

\textbf{Effect of the Metric Formulation during Training:}
When uncertain, the proposed metric tends to treat the pair similarly since we think an uncertain metric should not be able to differentiate all pairs. 
The proposed introspective similarity metric (ISM-Sim) based on the cosine similarity is defined as follows:
\begin{eqnarray}\label{eq:sim}
{C}_{IN}(\mathbf{x}_i, \mathbf{p}_j)=1  -  (1-C(\mathbf{x}_i, \mathbf{p}_j))\cdot e^{(-\frac{1}{\tau} \  \text{r\_conf}(\mathbf{x}_i, \mathbf{p}_j))}.
\end{eqnarray}
Alternatively, we may also weaken the similarity judgment by encouraging the metric to output large distances to all uncertainty pairs.
As a comparison, we additionally modified the metric to treat each ambiguous pair dissimilar (ISM-Dis) as follows:
\begin{eqnarray}\label{eq:dis}
{C}_{IN}(\mathbf{x}_i, \mathbf{p}_j)= C(\mathbf{x}_i, \mathbf{p}_j)\cdot e^{(-\frac{1}{\tau} \  \text{r\_conf}(\mathbf{x}_i, \mathbf{p}_j))}.
\end{eqnarray}
We conducted experiments on the CUB-200-2011 and Cars196 datasets to test the performances of using different metric formulations for training in Table~\ref{tab:dissimilar}. 
We observe that treating each ambiguous pair dissimilar performs worse than the original metric.
This verifies our motivation for using uncertainty to weaken the semantic discrepancy.

\textbf{Effect of the Metric Formulation during Testing:}
During testing, we adopt the original similarity metric without our uncertainty-aware term. 
As an alternative, we conducted an experiment using the introspective similarity metric (ISM) during testing, as shown in Table~\ref{tab:ISM}. 
We observe a decrease in performance when using ISM during testing, indicating a harmful effect of using uncertainty to weaken the similarity discrepancy during inference.
This is reasonable since providing a clear and confident similarity judgment is more beneficial to discriminative tasks.

\begin{table}[t] \small
\centering
\caption{Analysis of the batch size (BS) on the CUB-200-2011, Cars196, and Stanford Online Products datasets.}
\vspace{-7pt}
\label{tab:bs}
\setlength\tabcolsep{9pt}
\arraysep
\begin{tabular}{c|cc|cc|cc}
\hline
& \multicolumn{2}{|c}{CUB-200-2011} & \multicolumn{2}{|c}{Cars196} & \multicolumn{2}{|c}{SOP}\\
 \hline
BS & R@1 & NMI & R@1 & NMI & R@1 & NMI\\
\hline
40 & 66.3 & 70.6 & 88.2 & 74.1 & 78.1 & 89.5\\ 
60 & 67.1 & 71.3 & 88.9 & 75.3 & 79.9 & 91.3\\
80 & 68.5 & 72.0 & 89.6 & 75.6 & 80.5 & 91.6\\
100 & 69.6 & 72.5 & 90.0 & 76.3 & 81.2 & 92.1\\
120 & \textbf{70.7} & 73.5 & 90.6 & 76.9 & 81.5 & \textbf{92.3}\\
140 & 70.5 & 73.2 & 90.5 & 77.2 & \textbf{81.6} & 92.2\\
160 & \textbf{70.7} & 73.3 & \textbf{90.7} & 76.8 & 81.3 & \textbf{92.3}\\
180 & \textbf{70.7} & \textbf{73.6} & 90.5 & \textbf{77.5} & 81.3 & 92.1\\
\hline
\end{tabular}
\vspace{-2mm}
\end{table}

\textbf{Effect of the Batch Size:}
We conducted experiments to investigate the influence of different batch sizes during training. 
Specifically, we set the batch size from $40$ to $180$, as shown in Table~\ref{tab:bs}. 
We observe a relatively consistent performance improvement as the batch size increases.
This is because larger batch sizes enable richer relation mining among data.
Still, we see that the performance plateaus and even decreases when the batch sizes exceed $120$. Therefore, we set the batch size to 120 for the main experiments for a better balance of performance and computation.

\begin{figure}[t]
\begin{minipage}[t]{0.225\textwidth}
  \centering
\includegraphics[width=0.98\textwidth]{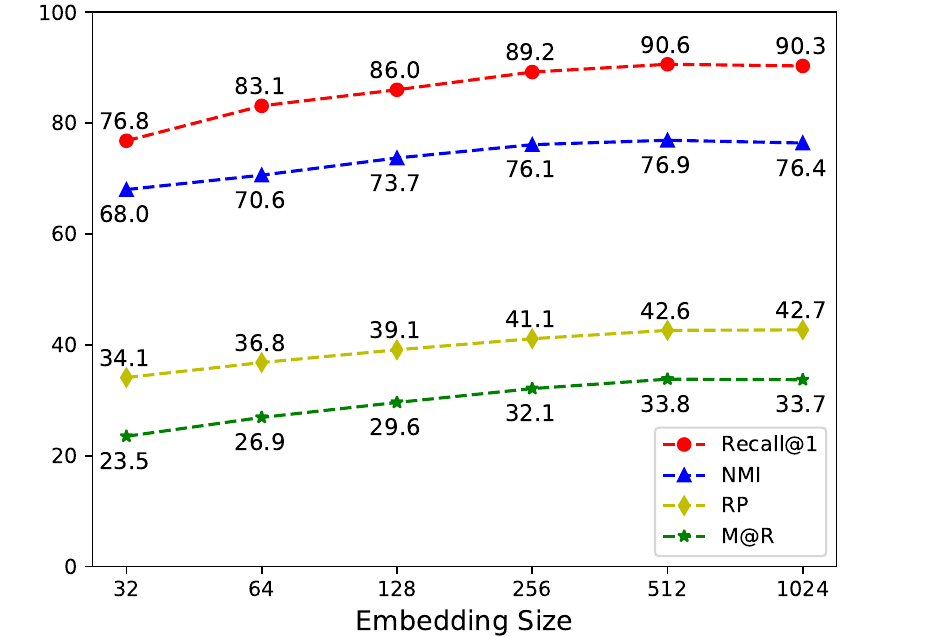}
\vspace{-7mm}
  \caption{
  Effect of the embedding size on Cars196.
  }
\vspace{-6mm}
\label{fig:es}
  \end{minipage}
~~~~
\begin{minipage}[t]{0.225\textwidth}
  \centering
\includegraphics[width=0.98\textwidth]{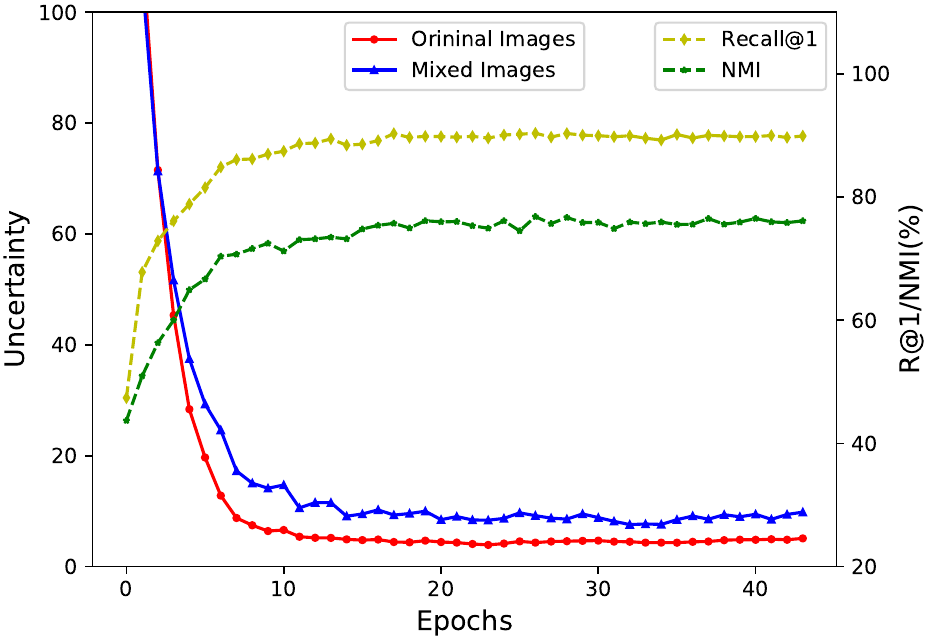}
\vspace{-7mm}
  \caption{
Uncertainty trend during training on Cars196.
}
\vspace{-6mm}
\label{fig:ue}
  \end{minipage}
\end{figure}

\textbf{Effect of the Embedding Size:}
We provided the experimental results with different embedding sizes on the Cars196 dataset in Figure~\ref{fig:es}. We observe that the performance already surpasses existing methods when the embedding size reaches $256$, indicating the superiority of the proposed IDML framework. 

\textbf{Uncertainty Trend during Training:}
To demonstrate that our IDML framework properly handles mixed images with high uncertainty, we visualize the trend of uncertainty level for both original images and mixed images during training, as shown in Figure~\ref{fig:ue}.
We define the uncertainty level of an image to be the L2-norm of its uncertainty embedding.
We see that the uncertainty decreases for both original and mixed images as the training proceeds and becomes stable as the model converges. 
We also observe that the uncertainty level for mixed images is larger than that of original images.
This verifies that our framework can indeed learn the uncertainty in images.

\begin{figure}[t]
\centering
\subcaptionbox{Effect of the triplet loss.\label{subfig:triplet}}{
\includegraphics[width=0.2\textwidth]{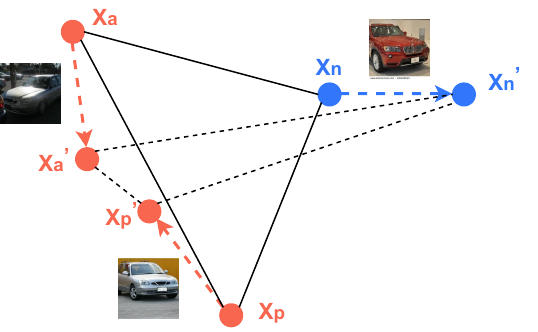}
}
\subcaptionbox{Effect of IDML-Triplet.\label{subfig:idml-triplet}}{
\includegraphics[width=0.175\textwidth]{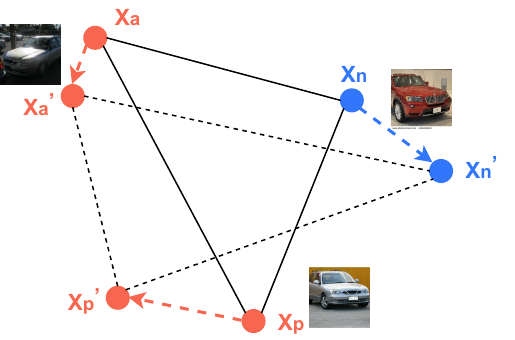}
}
\vspace{-3mm}
\caption{T-SNE~\cite{vandermaaten2014accelerating} analysis of one-step updating of the embeddings.}
\label{fig:qualitative results}
\vspace{-3mm}
\end{figure}

\textbf{Qualitative Analysis of the Learning Process:}
We provide a t-SNE~\cite{vandermaaten2014accelerating} visualized analysis of how embeddings are learned using a toy example on the Cars196 dataset in Figure~\ref{fig:qualitative results}. 
We visualized the embeddings before and after updating the model with one gradient step.
For a dark (and ambiguous) image $\mathbf{x}_\alpha$, the original method pulls it quite closer to the positive sample $\mathbf{x}_p$, while the proposed IDML is more cautious and only slightly pulls it together due to the uncertainty to prevent the influence of possible noise.

\begin{figure}[t] 
\centering
\includegraphics[width=0.45\textwidth]{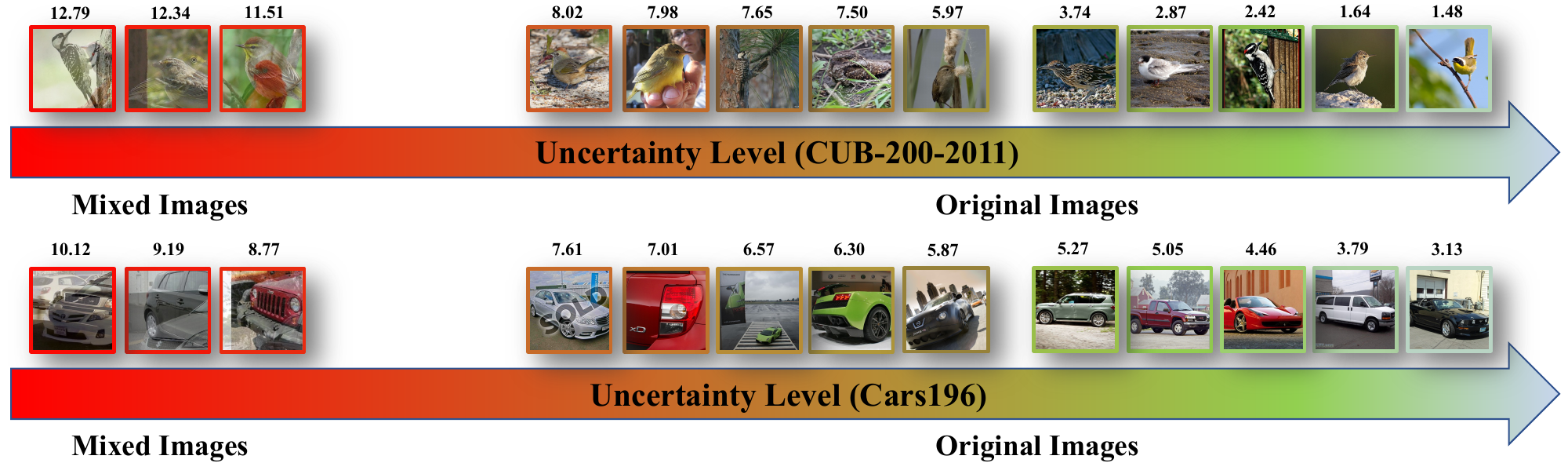}
\vspace{-3mm}
\caption{Uncertainty levels produced by the proposed IDML framework on the test split of CUB-200-2011 and Cars196.
} 
\label{fig:qr}
\vspace{-5mm}
\end{figure}

\textbf{Uncertainty Levels on the Test Split:}
We visualize the uncertainty levels on the test split of the CUB-200-2011 and Cars196 datasets, as shown in Figure~\ref{fig:qr}. 
We obtain the uncertainty levels of the mixed images together with original images in the test set after the model converges. 
We observe that the uncertainty of mixed images is much larger than that of the original test images since the mixed images contain the information of two images.
Also, we see that several original images result in relatively higher uncertainty than others because of the natural noise such as occlusion and improper directions. 
This further verifies that the proposed framework can successfully learn the uncertainty in images.

\section{Conclusion}
In this paper, we have presented an introspective deep metric learning (IDML) framework to properly process the data uncertainty for better performance. 
We have represented an image with a semantic embedding and an uncertainty embedding to model the semantic characteristics and the uncertainty, respectively.
We have further proposed an introspective similarity metric to compute an uncertainty-aware similarity score, which weakens semantic discrepancies for uncertain images. 
We have performed various experiments on the widely used benchmark datasets on both image retrieval to analyze the effectiveness of the proposed framework. 
Experimental results have demonstrated a constant performance boost to various methods in different settings.
It is interesting to apply our framework to self-supervised learning and vision transformers as future work.


{\small
\begin{thebibliography}{10}\itemsep=-1pt

\bibitem{babenko2014neural}
Artem Babenko, Anton Slesarev, Alexandr Chigorin, and Victor Lempitsky.
\newblock Neural codes for image retrieval.
\newblock In {\em ECCV}, pages 584--599, 2014.

\bibitem{cakir2019deep}
Fatih Cakir, Kun He, Xide Xia, Brian Kulis, and Stan Sclaroff.
\newblock Deep metric learning to rank.
\newblock In {\em CVPR}, pages 1861--1870, 2019.

\bibitem{chang2020data}
Jie Chang, Zhonghao Lan, Changmao Cheng, and Yichen Wei.
\newblock Data uncertainty learning in face recognition.
\newblock In {\em CVPR}, pages 5710--5719, 2020.

\bibitem{chen2020simple}
Ting Chen, Simon Kornblith, Mohammad Norouzi, and Geoffrey Hinton.
\newblock A simple framework for contrastive learning of visual representations.
\newblock In {\em ICML}, pages 1597--1607, 2020.

\bibitem{chen2021exploring}
Xinlei Chen and Kaiming He.
\newblock Exploring simple siamese representation learning.
\newblock In {\em CVPR}, pages 15750--15758, 2021.

\bibitem{chou2020remix}
Hsin-Ping Chou, Shih-Chieh Chang, Jia-Yu Pan, Wei Wei, and Da-Cheng Juan.
\newblock Remix: Rebalanced mixup.
\newblock In {\em ECCV}, pages 95--110, 2020.

\bibitem{chun2021probabilistic}
Sanghyuk Chun, Seong~Joon Oh, Rafael~Sampaio de Rezende, Yannis Kalantidis, and Diane Larlus.
\newblock Probabilistic embeddings for cross-modal retrieval.
\newblock In {\em CVPR}, pages 8415--8424, 2021.

\bibitem{deng2019arcface}
Jiankang Deng, Jia Guo, Niannan Xue, and Stefanos Zafeiriou.
\newblock Arcface: Additive angular margin loss for deep face recognition.
\newblock In {\em CVPR}, pages 4690--4699, 2019.

\bibitem{do2019theoretically}
Thanh-Toan Do, Toan Tran, Ian Reid, Vijay Kumar, Tuan Hoang, and Gustavo Carneiro.
\newblock A theoretically sound upper bound on the triplet loss for improving the efficiency of deep distance metric learning.
\newblock In {\em CVPR}, pages 10404--10413, 2019.

\bibitem{duan2018deep}
Yueqi Duan, Wenzhao Zheng, Xudong Lin, Jiwen Lu, and Jie Zhou.
\newblock Deep adversarial metric learning.
\newblock In {\em CVPR}, pages 2780--2789, 2018.

\bibitem{grill2020bootstrap}
Jean-Bastien Grill, Florian Strub, Florent Altch{\'e}, Corentin Tallec, Pierre Richemond, Elena Buchatskaya, Carl Doersch, Bernardo Avila~Pires, Zhaohan Guo, Mohammad Gheshlaghi~Azar, et~al.
\newblock Bootstrap your own latent-a new approach to self-supervised learning.
\newblock In {\em NeurIPS}, pages 21271--21284, 2020.

\bibitem{guo2020density}
Senhui Guo, Jing Xu, Dapeng Chen, Chao Zhang, Xiaogang Wang, and Rui Zhao.
\newblock Density-aware feature embedding for face clustering.
\newblock In {\em CVPR}, pages 6698--6706, 2020.

\bibitem{harwood2017smart}
Ben Harwood, Vijay Kumar B~G, Gustavo Carneiro, Ian Reid, and Tom Drummond.
\newblock Smart mining for deep metric learning.
\newblock In {\em ICCV}, pages 2840--2848, 2017.

\bibitem{he2020momentum}
Kaiming He, Haoqi Fan, Yuxin Wu, Saining Xie, and Ross Girshick.
\newblock Momentum contrast for unsupervised visual representation learning.
\newblock In {\em CVPR}, pages 9729--9738, 2020.

\bibitem{he2016deep}
Kaiming He, Xiangyu Zhang, Shaoqing Ren, and Jian Sun.
\newblock Deep residual learning for image recognition.
\newblock In {\em CVPR}, pages 770--778, 2016.

\bibitem{hershey2007approximating}
John~R Hershey and Peder~A Olsen.
\newblock Approximating the kullback leibler divergence between gaussian mixture models.
\newblock In {\em ICASSP}, pages IV--317, 2007.

\bibitem{hinton2015distilling}
Geoffrey Hinton, Oriol Vinyals, Jeff Dean, et~al.
\newblock Distilling the knowledge in a neural network.
\newblock In {\em NeurIPSW}, 2015.

\bibitem{hu2014discriminative}
Junlin Hu, Jiwen Lu, and Yap-Peng Tan.
\newblock Discriminative deep metric learning for face verification in the wild.
\newblock In {\em CVPR}, pages 1875--1882, 2014.

\bibitem{ioffe2015batch}
Sergey Ioffe and Christian Szegedy.
\newblock Batch normalization: Accelerating deep network training by reducing internal covariate shift.
\newblock In {\em ICLR}, pages 448--456, 2015.

\bibitem{kemertas2020rankmi}
Mete Kemertas, Leila Pishdad, Konstantinos~G Derpanis, and Afsaneh Fazly.
\newblock Rankmi: A mutual information maximizing ranking loss.
\newblock In {\em CVPR}, pages 14362--14371, 2020.

\bibitem{kendall2017uncertainties}
Alex Kendall and Yarin Gal.
\newblock What uncertainties do we need in bayesian deep learning for computer vision?
\newblock In {\em NeurIPS}, pages 5574--5584, 2017.

\bibitem{kim2020proxy}
Sungyeon Kim, Dongwon Kim, Minsu Cho, and Suha Kwak.
\newblock Proxy anchor loss for deep metric learning.
\newblock In {\em CVPR}, pages 3238--3247, 2020.

\bibitem{kim2018attention}
Wonsik Kim, Bhavya Goyal, Kunal Chawla, Jungmin Lee, and Keunjoo Kwon.
\newblock Attention-based ensemble for deep metric learning.
\newblock In {\em ECCV}, pages 760--777, 2018.

\bibitem{krause20133d}
Jonathan Krause, Michael Stark, Jia Deng, and Li Fei-Fei.
\newblock 3d object representations for fine-grained categorization.
\newblock In {\em ICCVW}, pages 554--561, 2013.

\bibitem{li2021learning}
Wanhua Li, Xiaoke Huang, Jiwen Lu, Jianjiang Feng, and Jie Zhou.
\newblock Learning probabilistic ordinal embeddings for uncertainty-aware regression.
\newblock In {\em CVPR}, pages 13896--13905, 2021.

\bibitem{milbich2020diva}
Timo Milbich, Karsten Roth, Homanga Bharadhwaj, Samarth Sinha, Yoshua Bengio, Bj{\"o}rn Ommer, and Joseph~Paul Cohen.
\newblock Diva: Diverse visual feature aggregation fordeep metric learning.
\newblock In {\em ECCV}, pages 590--607, 2020.

\bibitem{movshovitz2017no}
Yair Movshovitz-Attias, Alexander Toshev, Thomas~K. Leung, Sergey Ioffe, and Saurabh Singh.
\newblock No fuss distance metric learning using proxies.
\newblock In {\em ICCV}, pages 360--368, 2017.

\bibitem{musgrave2020metric}
Kevin Musgrave, Serge Belongie, and Ser-Nam Lim.
\newblock A metric learning reality check.
\newblock In {\em ECCV}, 2020.

\bibitem{neelakantan2014efficient}
Arvind Neelakantan, Jeevan Shankar, Alexandre Passos, and Andrew McCallum.
\newblock Efficient non-parametric estimation of multiple embeddings per word in vector space.
\newblock In {\em EMNLP}, pages 1059--1069, 2014.

\bibitem{nguyen2017mixture}
Dai~Quoc Nguyen, Dat~Quoc Nguyen, Ashutosh Modi, Stefan Thater, and Manfred Pinkal.
\newblock A mixture model for learning multi-sense word embeddings.
\newblock {\em arXiv}, abs/1706.05111, 2017.

\bibitem{nguyen2010cosine}
Hieu~V Nguyen and Li Bai.
\newblock Cosine similarity metric learning for face verification.
\newblock In {\em ACCV}, pages 709--720, 2010.

\bibitem{oh2018modeling}
Seong~Joon Oh, Kevin~P Murphy, Jiyan Pan, Joseph Roth, Florian Schroff, and Andrew~C Gallagher.
\newblock Modeling uncertainty with hedged instance embeddings.
\newblock In {\em ICLR}, 2018.

\bibitem{qian2019softtriple}
Qi Qian, Lei Shang, Baigui Sun, and Juhua Hu.
\newblock Softtriple loss: Deep metric learning without triplet sampling.
\newblock In {\em ICCV}, pages 6450--6458, 2019.

\bibitem{roth2019mic}
Karsten Roth, Biagio Brattoli, and Bjorn Ommer.
\newblock Mic: Mining interclass characteristics for improved metric learning.
\newblock In {\em ICCV}, pages 8000--8009, 2019.

\bibitem{roth2020pads}
Karsten Roth, Timo Milbich, and Bjorn Ommer.
\newblock Pads: Policy-adapted sampling for visual similarity learning.
\newblock In {\em CVPR}, pages 6568--6577, 2020.

\bibitem{roth2022non}
Karsten Roth, Oriol Vinyals, and Zeynep Akata.
\newblock Non-isotropy regularization for proxy-based deep metric learning.
\newblock In {\em CVPR}, pages 7420--7430, 2022.

\bibitem{russakovsky2015imagenet}
Olga Russakovsky, Jia Deng, Hao Su, Jonathan Krause, Sanjeev Satheesh, Sean Ma, Zhiheng Huang, Andrej Karpathy, Aditya Khosla, Michael Bernstein, et~al.
\newblock Imagenet large scale visual recognition challenge.
\newblock {\em IJCV}, 115(3):211--252, 2015.

\bibitem{sanakoyeu2019divide}
Artsiom Sanakoyeu, Vadim Tschernezki, Uta Buchler, and Bjorn Ommer.
\newblock Divide and conquer the embedding space for metric learning.
\newblock In {\em CVPR}, pages 471--480, 2019.

\bibitem{schroff2015facenet}
Florian Schroff, Dmitry Kalenichenko, and James Philbin.
\newblock Facenet: A unified embedding for face recognition and clustering.
\newblock In {\em CVPR}, pages 815--823, 2015.

\bibitem{shaw2002signal}
Gary Shaw and Dimitris Manolakis.
\newblock Signal processing for hyperspectral image exploitation.
\newblock {\em SPM}, 19(1):12--16, 2002.

\bibitem{shi2019probabilistic}
Yichun Shi and Anil~K Jain.
\newblock Probabilistic face embeddings.
\newblock In {\em ICCV}, pages 6902--6911, 2019.

\bibitem{simonyan2014very}
Karen Simonyan and Andrew Zisserman.
\newblock Very deep convolutional networks for large-scale image recognition.
\newblock {\em arXiv}, abs/1409.1556, 2014.

\bibitem{song2016deep}
Hyun~Oh Song, Yu Xiang, Stefanie Jegelka, and Silvio Savarese.
\newblock Deep metric learning via lifted structured feature embedding.
\newblock In {\em CVPR}, pages 4004--4012, 2016.

\bibitem{sun2020view}
Jennifer~J Sun, Jiaping Zhao, Liang-Chieh Chen, Florian Schroff, Hartwig Adam, and Ting Liu.
\newblock View-invariant probabilistic embedding for human pose.
\newblock In {\em ECCV}, pages 53--70, 2020.

\bibitem{sun2020circle}
Yifan Sun, Changmao Cheng, Yuhan Zhang, Chi Zhang, Liang Zheng, Zhongdao Wang, and Yichen Wei.
\newblock Circle loss: A unified perspective of pair similarity optimization.
\newblock In {\em CVPR}, pages 6398--6407, 2020.

\bibitem{szegedy2015going}
Christian Szegedy, Wei Liu, Yangqing Jia, Pierre Sermanet, Scott~E Reed, Dragomir Anguelov, Dumitru Erhan, Vincent Vanhoucke, and Andrew Rabinovich.
\newblock Going deeper with convolutions.
\newblock In {\em CVPR}, pages 1--9, 2015.

\bibitem{teh2020proxynca++}
Eu~Wern Teh, Terrance DeVries, and Graham~W Taylor.
\newblock Proxynca++: Revisiting and revitalizing proxy neighborhood component analysis.
\newblock In {\em ECCV}, 2020.

\bibitem{vandermaaten2014accelerating}
Laurens Van Der~Maaten.
\newblock Accelerating t-sne using tree-based algorithms.
\newblock {\em JMLR}, 15(1):3221--3245, 2014.

\bibitem{verma2019manifold}
Vikas Verma, Alex Lamb, Christopher Beckham, Amir Najafi, Ioannis Mitliagkas, David Lopez-Paz, and Yoshua Bengio.
\newblock Manifold mixup: Better representations by interpolating hidden states.
\newblock In {\em ICML}, pages 6438--6447, 2019.

\bibitem{vilnis2015word}
Luke Vilnis and Andrew McCallum.
\newblock Word representations via gaussian embedding.
\newblock In {\em ICLR}, 2015.

\bibitem{wah2011caltech}
Catherine Wah, Steve Branson, Peter Welinder, Pietro Perona, and Serge~J Belongie.
\newblock The {Caltech-UCSD Birds-200-2011} dataset.
\newblock Technical Report CNS-TR-2011-001, California Institute of Technology, 2011.

\bibitem{wang2018cosface}
Hao Wang, Yitong Wang, Zheng Zhou, Xing Ji, Dihong Gong, Jingchao Zhou, Zhifeng Li, and Wei Liu.
\newblock Cosface: Large margin cosine loss for deep face recognition.
\newblock In {\em CVPR}, pages 5265--5274, 2018.

\bibitem{wang2014learning}
Jiang Wang, Yang Song, Thomas Leung, Chuck Rosenberg, Jingbin Wang, James Philbin, Bo Chen, and Ying Wu.
\newblock Learning fine-grained image similarity with deep ranking.
\newblock In {\em CVPR}, pages 1386--1393, 2014.

\bibitem{wang2017deep}
Jian Wang, Feng Zhou, Shilei Wen, Xiao Liu, and Yuanqing Lin.
\newblock Deep metric learning with angular loss.
\newblock In {\em ICCV}, pages 2593--2601, 2017.

\bibitem{wang2019multi}
Xun Wang, Xintong Han, Weilin Huang, Dengke Dong, and Matthew~R Scott.
\newblock Multi-similarity loss with general pair weighting for deep metric learning.
\newblock In {\em CVPR}, pages 5022--5030, 2019.

\bibitem{wang2019ranked}
Xinshao Wang, Yang Hua, Elyor Kodirov, Guosheng Hu, Romain Garnier, and Neil~M Robertson.
\newblock Ranked list loss for deep metric learning.
\newblock In {\em CVPR}, pages 5207--5216, 2019.

\bibitem{wang2020cross}
Xun Wang, Haozhi Zhang, Weilin Huang, and Matthew~R Scott.
\newblock Cross-batch memory for embedding learning.
\newblock In {\em CVPR}, pages 6388--6397, 2020.

\bibitem{wu2017sampling}
Chao-Yuan Wu, R Manmatha, Alexander~J Smola, and Philipp Kr{\"a}henb{\"u}hl.
\newblock Sampling matters in deep embedding learning.
\newblock In {\em ICCV}, pages 2859--2867, 2017.

\bibitem{xuan2018deep}
Hong Xuan, Richard Souvenir, and Robert Pless.
\newblock Deep randomized ensembles for metric learning.
\newblock In {\em ECCV}, pages 723--734, 2018.

\bibitem{yang2019learning}
Lei Yang, Xiaohang Zhan, Dapeng Chen, Junjie Yan, Chen~Change Loy, and Dahua Lin.
\newblock Learning to cluster faces on an affinity graph.
\newblock In {\em CVPR}, pages 2298--2306, 2019.

\bibitem{ye2020probabilistic}
Mang Ye and Jianbing Shen.
\newblock Probabilistic structural latent representation for unsupervised embedding.
\newblock In {\em CVPR}, pages 5457--5466, 2020.

\bibitem{yu2019deep}
Baosheng Yu and Dacheng Tao.
\newblock Deep metric learning with tuplet margin loss.
\newblock In {\em ICCV}, pages 6490--6499, 2019.

\bibitem{yu2013kl}
Dong Yu, Kaisheng Yao, Hang Su, Gang Li, and Frank Seide.
\newblock Kl-divergence regularized deep neural network adaptation for improved large vocabulary speech recognition.
\newblock In {\em ICASSP}, pages 7893--7897, 2013.

\bibitem{yuan2019signal}
Tongtong Yuan, Weihong Deng, Jian Tang, Yinan Tang, and Binghui Chen.
\newblock Signal-to-noise ratio: A robust distance metric for deep metric learning.
\newblock In {\em CVPR}, pages 4815--4824, 2019.

\bibitem{yuan2017hard}
Yuhui Yuan, Kuiyuan Yang, and Chao Zhang.
\newblock Hard-aware deeply cascaded embedding.
\newblock In {\em ICCV}, pages 814--823, 2017.

\bibitem{yun2019cutmix}
Sangdoo Yun, Dongyoon Han, Seong~Joon Oh, Sanghyuk Chun, Junsuk Choe, and Youngjoon Yoo.
\newblock Cutmix: Regularization strategy to train strong classifiers with localizable features.
\newblock In {\em ICCV}, pages 6023--6032, 2019.

\bibitem{zhang2021point}
Biao Zhang and Peter Wonka.
\newblock Point cloud instance segmentation using probabilistic embeddings.
\newblock In {\em CVPR}, pages 8883--8892, 2021.

\bibitem{zhang2022attributable}
Borui Zhang, Wenzhao Zheng, Jie Zhou, and Jiwen Lu.
\newblock Attributable visual similarity learning.
\newblock In {\em CVPR}, pages 7532--7541, 2022.

\bibitem{zhang2018mixup}
Hongyi Zhang, Moustapha Cisse, Yann~N Dauphin, and David Lopez-Paz.
\newblock mixup: Beyond empirical risk minimization.
\newblock In {\em ICLR}, 2018.

\bibitem{zhang2017learning}
Li Zhang, Tao Xiang, and Shaogang Gong.
\newblock Learning a deep embedding model for zero-shot learning.
\newblock In {\em CVPR}, pages 2021--2030, 2017.

\bibitem{zhao2021towards}
Wenliang Zhao, Yongming Rao, Ziyi Wang, Jiwen Lu, and Jie Zhou.
\newblock Towards interpretable deep metric learning with structural matching.
\newblock In {\em CVPR}, pages 9887--9896, 2021.

\bibitem{zheng2019hardness}
Wenzhao Zheng, Zhaodong Chen, Jiwen Lu, and Jie Zhou.
\newblock Hardness-aware deep metric learning.
\newblock In {\em CVPR}, pages 72--81, 2019.

\bibitem{zheng2020structural}
Wenzhao Zheng, Jiwen Lu, and Zhou Jie.
\newblock Structural deep metric learning for room layout estimation.
\newblock In {\em ECCV}, 2020.

\bibitem{zheng2020deep}
Wenzhao Zheng, Jiwen Lu, and Jie Zhou.
\newblock Deep metric learning via adaptive learnable assessment.
\newblock In {\em CVPR}, pages 2960--2969, 2020.

\bibitem{zheng2021hardness}
Wenzhao Zheng, Jiwen Lu, and Jie Zhou.
\newblock Hardness-aware deep metric learning.
\newblock {\em TPAMI}, 43(09):3214--3228, 2021.

\bibitem{zheng2021deep}
Wenzhao Zheng, Chengkun Wang, Jiwen Lu, and Jie Zhou.
\newblock Deep compositional metric learning.
\newblock In {\em CVPR}, pages 9320--9329, 2021.

\bibitem{zheng2021deepr}
Wenzhao Zheng, Borui Zhang, Jiwen Lu, and Jie Zhou.
\newblock Deep relational metric learning.
\newblock In {\em ICCV}, pages 12065--12074, 2021.

\bibitem{zhu2020imbalance}
Hao Zhu, Yang Yuan, Guosheng Hu, Xiang Wu, and Neil Robertson.
\newblock Imbalance robust softmax for deep embedding learning.
\newblock In {\em ACCV}, 2020.

\end{thebibliography}

}
\end{document}